\documentclass[fleqn,10pt]{wlscirep}
\usepackage[utf8]{inputenc}
\usepackage[T1]{fontenc}

\usepackage{url}
\usepackage{graphicx}
\usepackage[table,xcdraw]{xcolor}

\usepackage[most]{tcolorbox}
\usepackage{xcolor}
\newtcolorbox{quotebox}{
  colback=gray!10,      
  colframe=gray!40,     
  boxrule=0.4pt,
  arc=2pt,
  left=6pt,right=6pt,top=6pt,bottom=6pt,
  left skip=2em,        
  right skip=2em        
}

\usepackage{lineno}

\title{AI agents reshape consensus formation in human groups}

\author[1,2]{Lin Chen}
\author[2,3]{Ziyi Liu}
\author[4*]{Xia Hu}
\author[2,3,*]{Yong Li}
\affil[1]{Network Science Institute, Northeastern University, Boston, MA, USA}
\affil[2]{Department of Electronic Engineering, Tsinghua University, Beijing, P.R. China}
\affil[3]{Beijing National Research Center for Information Science and Technology (BNRist), Tsinghua University, Beijing, P.R. China}
\affil[4]{Shanghai Artificial Intelligence Laboratory, Shanghai, P.R. China}
\affil[*]{Corresponding authors: xia.hu@rice.edu, liyong07@tsinghua.edu.cn}

\begin{abstract}

As large language model (LLM) agents shift from tools to participants in human groups, a fundamental question for collective behavior is how their growing presence reshapes consensus formation. 
Here we study mixed human–AI groups in a collaborative description game, in which shared conventions emerge through repeated rounds of random pairwise communication. 
Varying the proportions of LLM agents, we identify three distinct regimes of consensus formation: low agent proportions facilitate human-led consensus, intermediate proportions disrupt convergence, and high proportions restore strong consensus while shifting it toward agent-led conventions.
Crucially, these regimes differ not only in the strength of convergence, but also in the semantic grounding and communicative form of the resulting consensus: human-led consensus is more concrete, holistic, and grounded in shared real-world analogies, whereas agent-led consensus is more abstract, less information-dense, and more geometrically segmented. 
Mechanistically, agent influence arises from a shared linguistic prior that places agents near one another in the expression space, combined with relatively stable expression choices across rounds; humans initially resist adopting expressions from partners perceived as AI but gradually yield to conformity pressure. 
These findings provide evidence that AI composition can shape the emergence, content, and perceived legitimacy of group norms, making agent proportion and transparency important design variables for human–AI systems.

\end{abstract}

\begin{document}

\flushbottom
\maketitle
%
%
\thispagestyle{empty}

\section*{Introduction}

Large language model (LLM) agents are increasingly embedded in human social environments, including workplaces, online communities, and collaborative platforms~\cite{vaccaro2024combinations,bae2021social,ta2020user}.
With rapidly developing capacities to incorporate interaction history and generate adaptive responses~\cite{park2023generative, strachan2024testing,tessler2024ai}, agents are becoming not only tools for individual users but also social interlocutors within group interaction~\cite{chen2026ai,teixeira2026human}.
A central process in such interaction is consensus formation, through which groups gradually converge on shared ways of communicating and acting~\cite{centola2015spontaneous}.
The entry of LLM agents into this process raises a fundamental question for the behavioral sciences: when artificial agents participate in human groups, do they simply amplify existing human-driven dynamics, or do they reshape the consensus that groups produce~\cite{gelfand2024norm}?
%

Prior work has shown that consensus can arise spontaneously in homogeneous populations, both in pure-human groups~\cite{centola2015spontaneous} and, more recently, in populations composed entirely of LLM agents~\cite{ashery2025emergent}.
However, real-world deployments generally fall between these extremes, involving hybrid groups in which humans and LLM agents, with potentially different behavioral tendencies and adaptation strategies, interact repeatedly and jointly shape group outcomes. 
Whether such mixed human–AI groups can form shared conventions—and whether the resulting outcome resembles either pure-human or pure-agent consensus—remains largely unknown. 
Existing research has provided evidence that AI can influence human judgment~\cite{glickman2025human}, persuasion~\cite{bai2025llm}, cooperation~\cite{fu2026optimal}, and team performance~\cite{vaccaro2024combinations}, but has primarily studied AI's impact on externally defined task outcomes rather than on emergent norms within a group.
Understanding these dynamics is critical not only for the practical design of human–AI systems that preserve human agency while leveraging AI capabilities, but also for developing theories of how machines mediate social and cultural processes~\cite{brinkmann2023machine,tsvetkova2024new}. 

Consensus in mixed human–AI groups may differ from its pure-human counterpart not only in strength, but also in whose expressions shape the emerging convention and what kind of meaning it carries.
Moreover, these dimensions need not move together: a group may converge strongly, yet the resulting convention may be disproportionately shaped by a subset of participants or carry qualitatively different meaning~\cite{xie2011social,ashery2025emergent}.
Such multidimensional shifts are difficult to isolate in field settings, where institutional roles, prior relationships, and communication histories are often confounded~\cite{speer2024hyperscanning,glickman2025human}.
We therefore adopt a collaborative description game based on the referential communication paradigm~\cite{clark1986referring,centola2015spontaneous,ashery2025emergent}, in which participants are paired at random and must coordinate on a shared description for a single ambiguous visual referent, through written exchanges with no externally imposed vocabulary.
Group-level conventions thereby arise not through centralized instruction, but through the accumulation of repeated local adjustments across many such pairings.
This paradigm provides a tractable model of how consensus emerges from distributed interaction: anonymous random pairing allows participants to interact without knowledge of their partner's identity, removing confounds from fixed social roles and prior relationships; the absence of an imposed vocabulary allows semantic content to emerge from participants' own expressions rather than from predefined options. 

Using this paradigm, we experimentally vary the proportion of LLM agents in repeated human–AI description games. 
We find that mixed human–AI groups do not change monotonically as agent proportion increases. Instead, they pass through three distinct regimes. 
At low proportions, agents facilitate and catalyze human consensus by adapting to human-like conventions; at intermediate proportions, convergence destabilizes as neither group fully adapts to the other; and at high proportions, agents restore strong consensus while redirecting it toward agent-authored norms. 
Moreover, as agent proportion increases, agents increasingly shape not only the lexical form but also the conceptual structure of the resulting consensus. 
This reshaping is visible in the content itself: while human groups produce concrete, information-dense descriptions built on real-world analogies, holistic forms, and dynamic narration, agent-led consensus yields more abstract, sparser accounts organized around geometric parts and static enumeration. 
Mechanistic analyses point to two coupled processes: on the agent side, a shared linguistic prior places agents close together in expression space from the outset, and their relatively stable expressions across rounds sustain this concentration, creating a semantic attractor that channels group convergence toward agent-authored conventions. 
On the human side, humans initially resist adopting expressions perceived as AI-generated, but as agent expressions converge near the group norm, conformity pressure increasingly overrides this resistance.
Together, these findings show that mixed human–AI collectives are a distinct class of social systems in which increasing AI participation reorganizes consensus not only in strength, but also in ownership and meaning.

\section*{Results}

\subsection*{Varying agent proportion reveals three regimes of consensus formation}

To investigate how LLM agents influence human consensus formation, we use a collaborative description game (Figure~\ref{fig1}a; see Supplementary Figure 1 for the interface of our experimental platform). 
In each round, participants are randomly paired and independently describe the same abstract shape (tangram). 
After submission, both players receive their partner's description along with a similarity score, providing feedback on their alignment. 
Following prior work~\cite{centola2015spontaneous}, this process repeats for 40 rounds, allowing participants to iteratively develop shared conventions for describing the shape. 
We systematically vary the proportion of LLM agents in the group across five conditions: 0\% (pure-human baseline), 12.5\%, 33.3\%, 50\%, and 75\%. 
Consensus strength is quantified as the average semantic similarity among participants' final-round descriptions (see Methods M1).

The relationship between agent proportion and consensus strength is non-monotonic, revealing three distinct regimes (Figure~\ref{fig1}b). 
Compared with the pure-human baseline group (mean consensus strength = 0.695), consensus strength at the low agent proportion (12.5\%) increases by 8.0\% on average ($z=3.282$, $p=0.001$).
However, at the intermediate proportions (33.3\%, 50\%), consensus formation is significantly impaired, dropping by 23.1\% ($z=-12.326$, $p<0.001$) and 14.5\% ($z=-6.156$, $p<0.001$), respectively.
Strikingly, at the high agent proportion (75\%), consensus strength rebounds to $0.725$, comparable to the level observed in the 12.5\% condition.
This non-monotonic pattern—catalysis at low proportions, impairment at intermediate proportions, and re-establishment at high proportions—suggests that agent influence may operate differently across these regimes.
We refer to these as H1 (low proportion, 12.5\%), H2 (intermediate proportions, 33.3\% and 50\%), and H3 (high proportion, 75\%) throughout the remainder of the paper.

We further analyze the directional movement of humans and agents toward each other's linguistic space (Figure~\ref{fig1}c).
We quantify this by measuring, for each individual, how much closer they move to the opposing group's initial linguistic position by the end of the experiment (see Methods M1).
At 12.5\%, agents tended to move more toward the human linguistic space than humans moved toward the agent space (average movement 0.822 for agents and 0.575 for humans, Mann–Whitney $U=18$, $p=0.135$), though this does not reach significance given the small number of agents in this condition ($n=3$). 
However, starting from 33.3\%, this asymmetry reverses: humans move toward agents significantly more than agents move toward humans (33.3\%: $U=104$, $p=0.007$; 50\%: $U=129$, $p<0.001$; 75\%: $U=103$, $p<0.001$), indicating that agent-led assimilation becomes the dominant mode of linguistic convergence as agent proportion increases.

This systematic reversal in convergence directionality explains the three-regime pattern of consensus formation. At low proportions, agents catalyze human-led consensus by adapting to facilitate convergence within the existing human linguistic space. At intermediate proportions, the competing directional forces—with neither group fully adapting to the other—create interference that disrupts consensus formation. At high proportions, the convergence direction has fully reversed, with humans now adapting toward the agents' linguistic space to establish a new consensus. Thus, although groups with low and high agent proportion achieved similar consensus strength, the former represents agent-facilitated human consensus, while the latter represents agent-led consensus—a mechanistic distinction that raises questions about whether these pathways also produce different consensus content, which we examine in the following section.




\subsection*{Agent participation transforms the semantic content of consensus}

The preceding analysis reveals that 12.5\% and 75\% agent conditions achieve similar consensus strength through opposing mechanisms: agent-facilitated human consensus versus agent-led consensus. This raises the question of whether these different pathways also produce different consensus content. 
To answer this question, we compare the final consensus formed under these two conditions, focusing on agent influence at lexical and conceptual levels, and on the semantic properties of the resulting expressions.

We first examine agent influence at two complementary levels: lexical contribution quantifies the degree of agent contribution to the final stable vocabulary, while conceptual contribution measures the degree of agent contribution to the semantic clusters (see Methods M2).
Across agent proportions, these two metrics exhibit different trajectories (Figure~\ref{fig2}a; Supplementary Figures 2-3). 
At the lexical level, agents' contribution substantially exceeds their proportion in the population, and the contribution rises monotonically.
Specifically, 33.3\% of agents already contribute 60.7\% of the final words. 
In contrast, at the conceptual level, agents' contribution exhibits threshold dynamics, rising slowly in the early stages and then rapidly dominating at high proportions, where it reaches 100\% contribution at the 75\% proportion, indicating complete agent dominance of the semantic space. 
This dissociation suggests that agents first shape the surface vocabulary before influencing the conceptual organization of consensus.
To quantify this shift from lexical to conceptual influence, we introduce the metric of \textit{conceptual-lexical ratio (CLR)}, defined as the ratio of conceptual-level to lexical-level agent contribution (Figure~\ref{fig2}b; see Methods M2).
$CLR > 1$ indicates that agents exert greater influence on conceptual content than their lexical footprint alone would suggest, consistent with a form of conceptual leadership, in which agents shape the semantic content of consensus rather than only its wording. 
$CLR < 1$ indicates the opposite: agents function primarily as "linguistic anchors," standardizing expression without commensurately redirecting semantic content.
At lower proportions (12.5\% and 33.3\%), CLR remains below 1, confirming that agents primarily provide lexical standardization. 
Starting from 50\%, CLR exceeds 1 (reaching 1.24 at 50\% and 1.27 at 75\%), indicating that agents shift from primarily lexical influence to stronger conceptual influence over the consensus.
This sequencing arises because lexical contribution requires only that agents introduce reusable words into the shared vocabulary, a process that can emerge even at low proportions because agents sharing a common pretrained prior tend to produce similar terms independently across interactions. Conceptual restructuring, by contrast, requires a critical mass of agents whose shared prior exerts sufficient sustained directional pressure to displace the human-derived semantic framework.

Having established that human-led (H1) and agent-led (H3) consensus differ in the depth of agent influence, we next examine whether they also differ in their content. 
We compare them along five dimensions—concreteness, propositional idea density, analogical ratio, holistic ratio, and event framing ratio—that capture the degree to which consensus content is concrete, information dense, analogical, holistic, and dynamically framed (see Methods M3; Figure~\ref{fig2}c). 
First, we examine \textit{concreteness}, which quantifies the degree to which words refer to tangible, perceptible entities that can be experienced through the senses~\cite{brysbaert2014concreteness}.
From human-led consensus to agent-led consensus, concreteness decreases significantly from 2.986 (95\% CI: [2.954, 3.017]) to 2.724 (95\% CI: [2.704, 2.745]; Mann–Whitney $U=12914$, $p<0.001$), indicating that agent-led consensus relies more on abstract descriptions.
Second, we measure \textit{propositional idea density}, defined as the number of distinct propositions or ideas expressed per unit of text, which reflects communicative efficiency~\cite{brown2008automatic}.
Propositional idea density declines from 5.419 (95\% CI: [5.240, 5.599]) to 5.321 (95\% CI: [5.230, 5.412]; $U=8448$, $p=0.020$), suggesting lower propositional density in agent-led consensus (Figure~\ref{fig2}d).
Third, \textit{analogical ratio} captures the proportion of expressions drawing on biological or object analogies from real-world experience. 
It decreases from 0.802 (95\% CI: [0.790, 0.814]) to 0.050 (95\% CI: [0.021, 0.078]; $U=14389$, $p<0.001$), showing reduced reliance on real-world analogies in agent-led consensus. 
Fourth, \textit{holistic ratio} measures the degree to which expressions describe whole-object forms rather than local geometric parts. 
It decreases from 0.685 (95\% CI: [0.670, 0.701]) to 0.353 (95\% CI: [0.339, 0.367]; $U=14182$, $p<0.001$), indicating a shift toward part-based geometric descriptions under agent-led consensus. 
Fifth, \textit{event framing ratio} captures the proportion of expressions that narrate actions or dynamic scenes rather than static enumerations. 
It decreases from 0.385 (95\% CI: [0.365, 0.404]) to 0.000 (95\% CI: [0.000, 0.000]; $U=14340$, $p<0.001$), reflecting diminished event-based narration in agent-led consensus.
The word clouds of stable consensus vocabulary crystallize this divergence visually (Figure~\ref{fig2}c, bottom): human-led consensus features embodied, scene-based vocabulary ("rabbit," "sit," "background"), whereas agent-led consensus is dominated by purely geometric enumeration ("triangle," "square," "asymmetrical").
Collectively, these patterns reveal a systematic transformation in consensus content: agent-led consensus shifts from concrete, information-dense, holistic descriptions grounded in real-world analogies and event-based narration toward abstract, information-sparse, part-based descriptions with less analogical grounding and less dynamic framing. 

This shift carries important implications for consensus quality and utility. 
Prior work on collaborative reference establishes that holistic analogies leverage shared real-world knowledge to build consensus, providing semantic depth beyond geometric description and enabling conversational partners to maintain perspective alignment across complex interactions without repeatedly re-parsing visual structure~\cite{clark1986referring}. 
Concreteness enhances memory encoding and subsequent reuse~\cite{fliessbach2006effect}. 
While part-based descriptions serve a valuable role in initial disambiguation~\cite{hupet1991effects}, they ultimately prove less efficient than compact holistic labels for sustained communication. 
Thus, although agent-led consensus achieves high strength, it produces descriptions that are harder to ground in shared experience, less informationally efficient, and more difficult for new members to adopt, even as the convergence itself masks these costs.




\subsection*{Initial alignment and adoption–persistence dynamics drive agent influence on consensus}
Having identified three regimes of agent influence on consensus strength and fundamental differences between agent- and human-led consensus content, we now examine the behavioral dynamics underlying these patterns.
We analyze what enables coordinated agent influence, how it operates through interaction dynamics, and whether manipulating agent behavioral traits alters consensus outcomes.
We first examine whether agents began the interaction from a more similar region of expression space than humans.
Before any interaction occurs, agent-agent initial similarity far exceeds human-human similarity across all agent proportions (Mann-Whitney $U=2214$, $p<0.001$; Figure~\ref{fig3}a). 
This baseline asymmetry reflects agents' shared pretrained linguistic prior, which concentrates them in the same region of expression space from the outset. 
This concentration creates a shared semantic anchor: agents began closer to one another than humans did, giving their expressions a common starting point from which coordinated influence could emerge.

We next analyze how participants adjust their expressions during the consensus formation process, because a shared starting point alone does not explain how agent influence is sustained across multiple rounds of interaction.
Cultural transmission theory suggests that conventions spread through a balance between two forces: a coordination bias toward adopting a partner's expressions and an egocentric bias toward retaining one's own~\cite{tamariz2014cultural}. We operationalize these as \textit{adoption}, the extent to which individuals incorporate their partner's expressions, and \textit{persistence}, the extent to which individuals maintain their own prior expressions (see Methods M4).
These two dimensions capture competing forces in convergence: the pull toward others' expressions versus adherence to one's own established conventions.
We divide the 40-round interaction into early (rounds 1-20) and late (rounds 21-40) phases and examine adoption and persistence dynamics across all conditions (Figure~\ref{fig3}b-c). 
Both humans and agents shifted from higher adoption and lower persistence in early rounds to lower adoption and higher persistence in later rounds (Human: $U=2437$, $p<0.001$; Agent: $U=1088$, $p=0.022$). 
This shift reflects the natural arc of consensus formation: initial exploration and mutual adaptation, followed by consolidation and commitment to established conventions. 
Meanwhile, a critical asymmetry emerges: in the early phase, humans adopt substantially more than agents ($U=1735$, $p<0.001$). 
Moreover, the magnitude of behavioral shift is significantly larger for humans than agents in both adoption decline ($\Delta$: -0.055 vs. -0.020; $U=671$, $p<0.001$) and persistence increase ($\Delta$: +0.143 vs. +0.059; $U=1702$, $p<0.001$).
These results indicate that humans showed larger behavioral adjustments than agents during consensus formation, while agents' behavior remains comparatively stable—an asymmetry that, when agent proportion is sufficient, channels convergence toward the region agents already occupy.

To test whether the balance between adoption and persistence causally influences consensus outcomes, we manipulate agent behavior through prompt engineering (see Methods M4; Figure~\ref{fig3}d-e).
We compare three conditions at the 33.3\% agent proportion: \textit{neutral} (no imposed behavioral tendency), \textit{high-adoption} (agents prompted to increase incorporation of partner expressions), and \textit{high-persistence} (agents prompted to maintain their own expressions more strongly). 
Strengthening adoption does not significantly alter consensus strength ($z=1.107$, $p=0.2684$) but reduces agent conceptual influence, with CLR decreasing by $42.8\%$ (from 0.599 to 0.342).
In contrast, strengthening persistence significantly impairs consensus formation, reducing the consensus level by $9.6\%$ compared to the neutral condition ($z=-3.289$, $p=0.001$), while simultaneously elevating the CLR by $117.8\%$ (from 0.599 to 1.304).
These manipulation experiments reveal that both consensus strength and agent influence are jointly determined by the balance between adoption and persistence. 
High-persistence agents, lacking the early adaptation phase, fail to establish common ground with humans, resulting in significantly lower consensus strength and impaired convergence, but their rigid position dramatically increases their conceptual dominance. 
High-adoption agents preserve consensus strength but cede conceptual influence over the final consensus due to insufficient persistence. 
Together, these results demonstrate that influential agents are neither "the more persistent the better" nor "the more accommodating the better," but rather depend on a balance between early adaptation and late consolidation. 
Early adoption enables agents to enter the human linguistic space and establish a lexical foothold; late persistence allows them to maintain and deepen their semantic territory. 
Disrupting this sequential balance undermines either consensus formation itself or the agent's capacity to shape its content.

Beyond the adoption–persistence balance, we examine whether the agent model's capability itself affects influence dynamics. 
We repeat the 33.3\% agent condition using a weaker base model (\texttt{Qwen2.5-VL-7B-Instruct}) instead of the original model (\texttt{Qwen2.5-VL-32B-Instruct}).
The weaker model substantially reduces agent influence: consensus level rises back from 0.534 to 0.725 ($z=14.143$, $p<0.001$), approaching the human-led condition (Figure~\ref{fig3}f). 
Moreover, CLR decreases by $22.9\%$ (from 0.599 to 0.462; Figure~\ref{fig3}g), indicating that the weaker model's influence is confined more to the lexical surface and less able to penetrate the conceptual level. 
This pattern arises because the weaker model lacks the capacity to maintain consistent expressions from round to round.
Without such consistency, the semantic attractor weakens substantially, leaving insufficient directional pull on human participants, who then converge more readily among themselves.
Sufficient behavioral capacity is thus a prerequisite for the semantic attractor to function—without it, agent proportion alone cannot generate directional influence.

Together, these results suggest that agent influence depends on both initial alignment among agents and the stability of their expressions across interactions. When either model capacity or the adoption–persistence balance is disrupted, the directional pull toward agent-led consensus weakens.




\subsection*{Perceived partner identity and conformity pressure jointly shape human adoption}

The preceding analyses rely on objective measurements—embedding similarity, movement direction, and conceptual-lexical ratios. 
We next examine how human participants perceived agent influence and whether these perceptions shaped adoption decisions.
To address these questions, we collect and analyze post-experiment questionnaire responses from human participants (see ``Experimental Procedure'' in Supplementary Information).
We first examine two subjective indicators across three agent proportion regimes: agreement with the group's final description and perceived extent of AI leadership (Figure~\ref{fig4}a). 
Agreement with the group's final description decreases from H1 to H3 (Mann-Whitney $U=110$, $p=0.004$), while perceived extent of AI leadership increases ($U=32$, $p=0.060$). 
This divergence reveals that at high agent proportions, humans converge on conventions with which they reported lower agreement: despite converging on a shared expression, participants increasingly perceive AI leadership in shaping the final description and express lower satisfaction with it.

We next examine whether perceived AI identity directly affects adoption decisions. 
In the post-experiment questionnaire, participants are presented with nine sampled interactions (in randomized order), each showing their own and their partner's expression from that round. 
For each interaction, participants judge whether their partner is a human or an AI and rate their willingness to adopt the partner's expression on a 5-point Likert scale, ranging from 1, “strongly unwilling,” to 5, “strongly willing.”
We compute the adoption willingness gap—the difference in willingness between partners judged as AI and partners judged as human—across three regimes (Figure~\ref{fig4}b). 
At H1 and H2, the gap is negative ($U=17764$, $p<0.001$): participants are less willing to adopt expressions from partners they perceive as AI, consistent with an identity-based resistance that functions as a subjective legitimacy filter. 
At H3, however, this gap narrows and even reverses ($\Delta$ = +0.884, $p=0.010$), indicating that the identity-based resistance attenuates at high agent proportions and humans become more accepting of AI-generated expressions.

To decompose the forces driving adoption decisions, we construct an ordinary least squares (OLS) regression model to predict adoption willingness with four predictors (see Methods M5; Figure~\ref{fig4}c; Detailed results in Supplementary Table 1).
Perceived AI identity exerts a significant negative effect ($\beta=-0.463$, 95\% CI=[-0.671, -0.254], $p<0.001$). 
Self–peer similarity ($\beta=+0.195$, 95\% CI=[0.044, 0.346], $p=0.012$) and peer-centroid proximity ($\beta=+0.229$, 95\% CI=[0.058, 0.400], $p=0.009$) both exert significant positive effects, while peer description length does not reach significance ($\beta=+0.091$, 95\% CI=[-0.024, 0.206], $p=0.120$). 
The two significant positive predictors capture proximity at different scales: self–peer similarity reflects dyadic alignment between the participant and their partner, whereas peer-centroid proximity reflects alignment between the partner and the broader group norm. 
The contrast between perceived AI identity and the two proximity effects thus reveals competing forces governing adoption: identity-based resistance suppresses uptake of expressions perceived as AI-generated, while proximity promotes it. 
The group-level channel is particularly consequential, because peer-centroid proximity operationalizes conformity pressure: a structural force that, unlike dyadic alignment, scales with agent proportion.
We further examine how the conformity pressure channel evolves across regimes.
The peer-centroid proximity gap widens from H1 (0.011) to H3 (0.078; Figure~\ref{fig4}d), consistent with the attractor's compressive effect on agent expressions, though this increase does not reach statistical significance ($\beta=+0.066$, $p=0.170$) given that only six human participants comprise an H3 group.
This pattern reflects the semantic attractor's compressive effect: as agent proportion increases, the attractor draws agent expressions closer to the group centroid, creating a structural conformity pressure that strengthens progressively across regimes.

Together, these subjective and objective findings illuminate the coupled mechanisms underlying agent-led consensus. 
Two forces compete in shaping human adoption decisions: identity-based resistance suppresses uptake of expressions perceived as AI-generated, while conformity pressure promotes uptake of expressions close to the group norm. 
As the agent proportion increases, the semantic attractor compresses the expression space, amplifying conformity pressure. 
At the same time, the identity-based resistance itself weakens under sustained agent exposure.
This combination produced a dissociation at high agent proportions, where humans become most willing to adopt AI-generated expressions precisely where they are least satisfied with the resulting consensus.
This dissociation indicates that agent-led consensus may arise less from persuasion or endorsement than from structural conformity to expressions that have become central within the group.

\section*{Discussion}

Our findings provide experimental evidence that LLM agents can reshape the emergence of shared conventions in human groups, through the accumulation of small, locally rational adaptations.
We show that varying agent proportions produces non-monotonic changes in consensus strength, content, and perceived ownership that are difficult to predict by studying individual human-agent exchanges in isolation.
This form of influence differs fundamentally from the mechanisms studied in prior work on human-AI interaction. 
Prior work has demonstrated that AI can influence human judgment through persuasive messaging~\cite{bai2025llm} and facilitate group consensus when acting as a designated mediator~\cite{tessler2024ai}.
In our experiment, agents hold no designated mediator or facilitator role.  
Unlike persuasion, which requires specially crafted content, and mediation, which requires a designated role, the agent influence we document emerges naturally from ordinary participation in repeated group interaction. 
What makes this process particularly consequential is that it unfolds through locally benign decisions: no individual interaction is coercive or strategically manipulative, yet collectively these interactions can shift whose expressions come to shape the convention.
As LLM agents become more embedded in humans' collaborative environments, the conditions for less visible forms of collective realignment may become more common, and thus more consequential.

Our findings carry direct implications for how hybrid human-AI systems should be designed and governed.
Importantly, AI participation in human groups should be considered not only as a means of improving productivity, but also as a force that can shape shared norms~\cite{farrell2025large,teixeira2026human}.
The non-monotonic relationship between agent proportion and consensus outcome indicates that different deployment goals may require different levels of agent participation.
Where the goal is to enhance coordination while preserving human authorship of shared norms, as in deliberation platforms, creative teams, or participatory decision-making, low agent proportions are most appropriate: agents catalyze convergence while adapting to the existing human linguistic space, accelerating agreement without redirecting its content, which echoes the facilitation effect of AI mediators~\cite{tessler2024ai} without requiring a designated role. 
Where the goal is instead to delay premature convergence and avoid groupthink, as in brainstorming scenarios, intermediate proportions may serve as a useful counterweight to sustain productive disagreement: the competing directional forces between humans and agents forestall the consolidation of any single perspective.
High agent proportions warrant particular caution. 
Groups converge strongly in this regime, but on expressions that are more abstract, less informationally dense, and less grounded in human experiential knowledge—and crucially, human resistance to AI-generated expressions weakens precisely here. From the outside, such consensus may appear similar to consensus that emerged without agent influence. This mechanism is relevant to concerns raised by Schroeder et al.~\cite{schroeder2026malicious} about malicious AI swarms, in which coordinated agent populations may create an appearance of grassroots consensus, steering collective discourse in ways that participants neither initiate nor detect. Our findings provide an experimental quantification of the conditions under which this becomes possible. A complementary design lever concerns transparency. The resistance humans show toward perceived AI-generated expressions is not merely a subjective preference—it has measurable downstream effects on adoption dynamics and, by extension, on whose conventions ultimately prevail. Systems that obscure agent identity do not simply withhold information; they actively suppress a protective human tendency whose effects our data make legible.

This study has several limitations. 
First, our paradigm instantiates convention formation through a referential communication game involving abstract visual stimuli, a design that affords experimental control but brackets features of naturalistic social environments, including content sensitivity, power asymmetries, and emotional investment. Generalization to richer communicative contexts therefore warrants caution. That said, the use of content-neutral stimuli is precisely what allows us to isolate the effect of agent proportion on consensus dynamics, independent of topic-specific confounds that would otherwise be difficult to disentangle. 
Second, following established convention formation paradigms~\cite{centola2015spontaneous}, our groups interact through random pairing, corresponding to a fully mixed interaction network. 
While this is the only topology shown to produce global consensus within experimentally accessible time scales, it does not capture the structured networks of real social environments, where communities, opinion leaders, and information bottlenecks shape interactions. 
In such settings, network structure could also serve as a deliberate instrument to regulate agent influence on consensus~\cite{stewart2019information}. 
Future work can explore how network design can be used to govern the balance between human and agent influence in hybrid groups.
Third, our experiments use a single model family at two capability levels. 
Its training pipeline (large-scale pretraining followed by supervised fine-tuning and alignment) is representative of current LLM families including LLaMA, GPT, and Gemini. 
The main findings we report arise from this shared training paradigm rather than from any Qwen-specific design choice.
The quantitative phase boundaries, however, may be model-dependent. 
Future work can systematically compare across model families to establish which phase boundaries are general and which shift with architecture, training data, or alignment strategy.

In sum, our findings call for extending the study of AI agents beyond individual model capabilities to their collective behavioral consequences in human social environments. As recent work has argued for an AI agent behavioral science that emphasizes systematic observation of how agents act, adapt, and interact in context~\cite{chen2026ai}, our study provides empirical evidence for why such a perspective is necessary: the effects of AI participation on collective norms emerge from group composition and repeated interaction, and cannot be anticipated from the properties of any individual model alone. Accounting for these collective dynamics will require treating agent proportion, transparency, and interaction structure as first-order design variables in hybrid human-AI systems, enabling communities to benefit from AI coordination while preserving meaningful control over the norms they form.

\section*{Methods}

\subsection*{M1. Measuring Consensus Dynamics}

We measure two complementary aspects of group dynamics: how strongly the group converges on a shared expression (consensus strength), and who drives that convergence (directional movement).

\paragraph{Consensus strength.}
To quantify how aligned participants' expressions are at the end of the experiment, we embed each description using \texttt{all-MiniLM-L6-v2}, a widely used sentence embedding model (mean pooling over token representations, followed by L2 normalization) and compute pairwise cosine similarities across all $\binom{N}{2}$ participant pairs, where $N = 24$ is the group size. 
We first compute $y_{i,t}$, participant $i$'s mean similarity to all other group members at round $t$:

\begin{equation}
    y_{i,t} = \frac{1}{N-1} \sum_{j \neq i} 
    \frac{\mathbf{e}_i^{(t)} \cdot \mathbf{e}_j^{(t)}}
    {\|\mathbf{e}_i^{(t)}\|\,\|\mathbf{e}_j^{(t)}\|}
\end{equation}

where $\mathbf{e}_i^{(t)} \in \mathbb{R}^d$ is the embedding of participant $i$'s expression at round $t$. 
Because participants may differ in stable idiosyncratic expression styles (e.g., habitual verbosity or phrasing), which would inflate or deflate their pairwise similarities irrespective of actual convergence, we apply a within-individual centering transform before aggregation. 
Group-level consensus at round $t$ is then defined as:

\begin{equation}
    C^{(t)} = \frac{1}{N}\sum_{i=1}^{N} 
    \left( y_{i,t} - \bar{y}_i + \bar{y}_{\text{run}} \right)
\end{equation}

where $\bar{y}_i = \frac{1}{T}\sum_t y_{i,t}$ is participant $i$'s personal mean across all $T$ rounds, and $\bar{y}_{\text{run}}$ is the grand mean across all participants and rounds within the same experimental run. 
This transform removes each participant's fixed similarity offset while preserving the overall scale of the measure. 
Final consensus strength is reported as $C^{(T)}$, the value at the last round.

\paragraph{Directional movement.}
Consensus strength alone does not reveal whether convergence was driven by humans adapting to agents or agents adapting to humans. 
To capture this, we measure how much each participant's expression moved toward the opposite group's linguistic space over the course of the experiment. 
For each group, we compute two reference centroids at round 1 before any adaptation has occurred, by averaging the embeddings of all human participants ($\mathbf{c}_H^{(1)}$) and all agent participants ($\mathbf{c}_A^{(1)}$), respectively. 
For participant $i$, directional movement toward the opposite group is then defined as the reduction in cosine distance to that group's initial centroid:

\begin{equation}
    \Delta d_i = d\!\left(\mathbf{e}_i^{(1)},\, \mathbf{c}_{\text{opp}}^{(1)}\right) 
    - d\!\left(\mathbf{e}_i^{(T)},\, \mathbf{c}_{\text{opp}}^{(1)}\right)
\end{equation}

where $d(\mathbf{u}, \mathbf{v}) = 1 - \frac{\mathbf{u} \cdot \mathbf{v}}{\|\mathbf{u}\|\|\mathbf{v}\|}$ is cosine distance, and $\mathbf{c}_{\text{opp}}^{(1)}$ is the round-1 centroid of the opposite group. 
A positive $\Delta d_i$ indicates that participant $i$ moves closer to the opposite group's initial linguistic space, i.e., adapts toward them.

\subsection*{M2. Measuring Contribution to Final Consensus}
To understand not just how much consensus forms, but whose expressions drive it, we measure agent contribution at two levels of abstraction: lexical-level (the surface vocabulary that participants converge on) and conceptual-level (the deeper semantic territories those words inhabit).

\paragraph{Lexical-level contribution.}
We first identify the stable consensus vocabulary as the set of words that appear in every one of the final 5 rounds, after lowercasing, stop-word removal, and lemmatization. 
For each word in this set, we locate its first round of appearance and compute agent and human usage counts within a short early-diffusion window (default: 3 rounds; results are robust to window sizes of 1–3 rounds; see Supplementary Figures 4-6).
To account for unequal group sizes, we normalize each group's raw count by its population share before computing a per-word contribution score ranging from 0 (entirely human-driven) to 1 (entirely agent-driven). 
Lexical-level agent contribution, $C_{\text{lex}}$, is the frequency-weighted average of these scores across the stable vocabulary, where the weight for each word is its total usage count in the final 5 rounds.

\paragraph{Conceptual-level contribution.}

Lexical-level attribution captures which specific words enter the consensus but may miss deeper conceptual contributions: an agent might redirect the semantic territory of consensus without introducing the specific words that ultimately stabilize. 
To capture this, we cluster all unique expressions produced across the entire experiment into semantic groups. 
Each expression is first parsed into a scene graph~\cite{schuster2015generating}, i.e., a structured representation comprising objects (nouns), attributes (adjectival modifiers), and relations (verbal predicates), using spaCy's dependency parsing function~\cite{spacy}. 
Pairwise similarity between expressions is then computed using a soft variant of the SPICE metric~\cite{anderson2016spice}, which calculates weighted F1 scores over the objects, attributes, and relations of each scene graph pair. 
We cluster the resulting distance matrix using HDBSCAN~\cite{mcinnes2017hdbscan}, which automatically determines the number of clusters. 
Clustering hyperparameters are optimized per condition using Optuna~\cite{akiba2019optuna}, maximizing the silhouette score. 
Each cluster represents a distinct semantic territory. 
We then apply the same early-diffusion, population-normalized scoring procedure as at the lexical level, yielding conceptual-level agent contribution $C_\text{con}$.
Formal definitions of both metrics are provided in the Supplementary Information (``Agent Contribution'' section).

%
%
%
%


\paragraph{Conceptual-Lexical Ratio (CLR).}
We introduce the Conceptual-Lexical Ratio (CLR) to summarize the relationship between these two levels:

\begin{equation}
    \text{CLR} = \frac{C_{\text{con}} + \varepsilon}{C_{\text{lex}} + \varepsilon},
\end{equation}

where $\varepsilon = 0.01$ is a smoothing term to avoid division by zero. 
$\text{CLR} > 1$ indicates that agents exert greater influence at the conceptual level than their lexical footprint alone would suggest; $\text{CLR} < 1$ indicates the converse.

\subsection*{M3. Analyzing Consensus Content} 
We analyze five dimensions of consensus content that together capture whether expressions are grounded in tangible, real-world experience (concreteness), informationally dense (propositional idea density), anchored in real-world analogies rather than purely geometric description (analogical ratio), holistic rather than part-based (holistic ratio), and dynamically framed rather than static (event framing ratio).

\paragraph{Concreteness.}
We quantify the degree to which words in a consensus expression refer to 
tangible, perceptible entities using the concreteness lexicon of Brysbaert et al.~\cite{brysbaert2014concreteness}, which provides mean concreteness ratings on a 1--5 scale for approximately 40,000 English word lemmas. 
For each expression, we scan words sequentially, attempting bigram matches before falling back to unigram lookup; words absent from the lexicon are excluded. 
The concreteness score of an expression is the mean rating across all matched tokens.

\paragraph{Propositional idea density.}
We measure propositional idea density as the average number of distinct propositions per ten words, reflecting communicative efficiency~\cite{brown2008automatic}. 
Following the part-of-speech approximation of the Computerized Propositional Idea Density Rater (CPIDR) proposed in the above paper, we use spaCy~\cite{spacy} for tagging and count as propositions all tokens tagged as verbs (excluding auxiliaries), adjectives, adverbs, prepositions, or conjunctions.

\paragraph{Analogical ratio, holistic ratio, and event framing ratio.} Analogical ratio, holistic ratio, and event framing ratio are obtained using an LLM-as-a-judge approach~\cite{li2024llms}, in which an LLM (\texttt{Qwen2.5-72B-Instruct}, temperature = 0) is prompted to rate each expression on three independent continuous scales from 0 to 1, following the theoretical framework of Clark \& Wilkes-Gibbs~\cite{clark1986referring}. 
Each unique expression in the final 5 rounds is annotated once; identical expressions share the same scores. Analogical ratio captures the extent to which a description maps the figure onto a recognizable real-world entity, creature, or scene (0 = purely geometric/abstract, 1 = vivid real-world entity mapping). Holistic ratio measures whether the description treats the figure as a unified whole rather than decomposing it into distinct component parts (0 = systematic part enumeration, 1 = fully holistic). Event framing ratio captures the degree to which the description frames the figure as engaged in dynamic action or posture, distinguishing agentive verbs (sitting, running) from compositional/structural verbs (composed of, arranged) that do not contribute to the score (0 = purely static, 1 = rich dynamic scene). The complete annotation prompt is provided in the Supplementary Information (``Agent Trait Prompts'' section). We also confirm the scoring consistency with other annotation models (see Supplementary Figure 7).

%

\subsection*{M4. Measuring and Manipulating Behavioral Mechanisms}
To understand the behavioral dynamics underlying consensus formation, we decompose each participant's interaction into two competing tendencies: the degree to which they incorporate their partner's expression (adoption), and the degree to which they maintain their own prior expression (persistence). 

\paragraph{Measuring adoption.}
Adoption quantifies how much a participant moves toward their partner's expression following an interaction. 
For participant $i$ in round $r$, we define adoption as the change in cosine similarity between participant $i$'s own expression and their partner's expression from round $r-1$ to round $r$:

\begin{equation}
    \text{adoption}(i, r) = \text{sim}\!\left(\mathbf{e}_i^{(r)},\, 
    \mathbf{e}_{\text{partner}}^{(r)}\right) - \text{sim}\!\left(\mathbf{e}_i^{(r-1)},\, 
    \mathbf{e}_{\text{partner}}^{(r)}\right)
\end{equation}

A positive value indicates that participant $i$'s expression became more similar to their partner's after the interaction; a negative value indicates divergence.

\paragraph{Measuring persistence.}
Persistence quantifies how stable a participant's expression is across consecutive rounds. For participant $i$ in round $r$, we define persistence as the cosine similarity between their own expressions in round $r$ and round $r+1$:

\begin{equation}
\text{persistence}(i, r) = \text{sim}\!\left(\mathbf{e}_i^{(r)},\, 
\mathbf{e}_i^{(r-1)}\right)
\end{equation}

A value close to 1 indicates that participant $i$ maintained their expression 
largely unchanged across rounds; a lower value indicates greater variation.

\paragraph{Prompt engineering of agent behavior traits.}
To test whether the temporal balance between adoption and persistence causally influences consensus outcomes, we manipulate agent behavior through prompt engineering at the system prompt level, holding agent proportion fixed at 33.3\%. 
All agents receive the same base instruction describing the game rules. 
The three conditions differ only in an additional behavioral tendency clause appended to the system prompt (see the full prompts in Supplementary Information):

\begin{itemize}
    \item \textbf{Neutral}: no behavioral tendency clause is added. Agents are instructed only to maximize their accumulated payoff through alignment with their partner.

    \item \textbf{High adoption}: agents are instructed that they have a strong preference for aligning with their partner, and should carefully observe their partner's previous expression and tend to reuse or imitate their phrasing and structure to increase similarity.

    \item \textbf{High persistence}: agents are instructed that they have a strong preference for maintaining a consistent and stable description style across rounds, should keep their wording, sentence structure, and choice of anchor words as similar as possible to their previous descriptions, and should avoid adopting expressions from Player 2 unless it matches their own interpretation of the image.
\end{itemize}

\subsection*{M5. Regression Model for Human Adoption Willingness}
To decompose the forces driving human adoption decisions, we fit an OLS regression predicting adoption willingness with heteroskedasticity-consistent standard errors (HC3). 
The model includes agent proportion condition as a categorical control and four predictors of interest: perceived AI identity (binary: 1 if the participant judged their partner as AI, 0 otherwise), self–peer similarity (cosine similarity between the participant's and their partner's expression embeddings), peer description length (word count of the partner's expression), and peer-centroid proximity (cosine similarity between the partner's expression embedding and the group centroid at that round, where the centroid is computed by averaging all participants' embeddings in the same round). All continuous predictors are z-score standardized; perceived AI identity is entered as a binary indicator.

\section*{Data Availability}
All data needed to evaluate the conclusions in the paper are present in the paper and/or the Supplementary Materials.

\section*{Code Availability}
The Python code to reproduce the main results in this paper is available in the GitHub repository: \url{https://github.com/tsinghua-fib-lab/human-ai-consensus}.

\bibliography{sample}


\section*{Author contributions}

Y.L., L.C., and X.H. conceived the project and the research outline. 
L.C. and Y.L. designed the research methods.
L.C. and Z.L. conducted the experiments.
L.C. prepared the figures.
All authors analyzed the results and participated in the writing of the manuscript.

\section*{Competing Interests}
The authors declare no competing interests.

\section*{Additional information}
Supplementary Information is available for this manuscript.

\clearpage

\begin{figure}[ht]
\centering
\includegraphics[width=\linewidth]{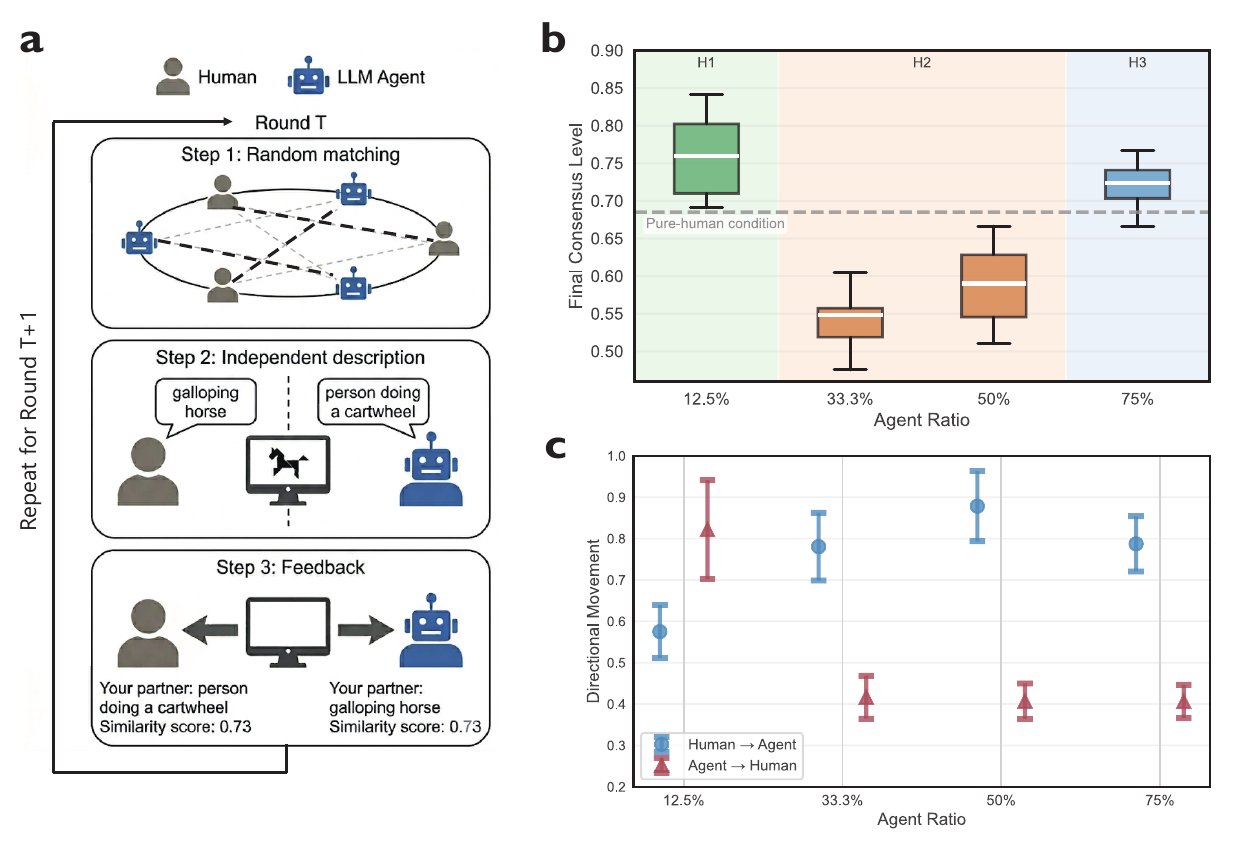}
\caption{\textbf{Experimental paradigm and phase-dependent consensus outcomes across agent proportions.} 
\textbf{a}, Schematic of the collaborative description game. Participants (humans and LLM agents) are assigned to groups and interact through repeated pairwise communication over 40 rounds. In each round, paired participants describe the same abstract tangram shape, receive the partner’s description and a similarity score, and proceed to the next round with new pairings. 
\textbf{b}, Final consensus strength as a function of agent proportion, quantified by semantic convergence of final-round expressions using embedding similarity. The dashed horizontal line marks the pure-human condition. Error bars indicate 95\% CI. 
\textbf{c}, Movement of humans toward the agent expression space (blue circles) and of agents toward the human expression space (red triangles) across agent proportions, measured in the embedding space. Error bars indicate 95\% CI. 
}
\label{fig1}
\end{figure}

\clearpage

\begin{figure}[ht]
\centering
\includegraphics[width=\linewidth]{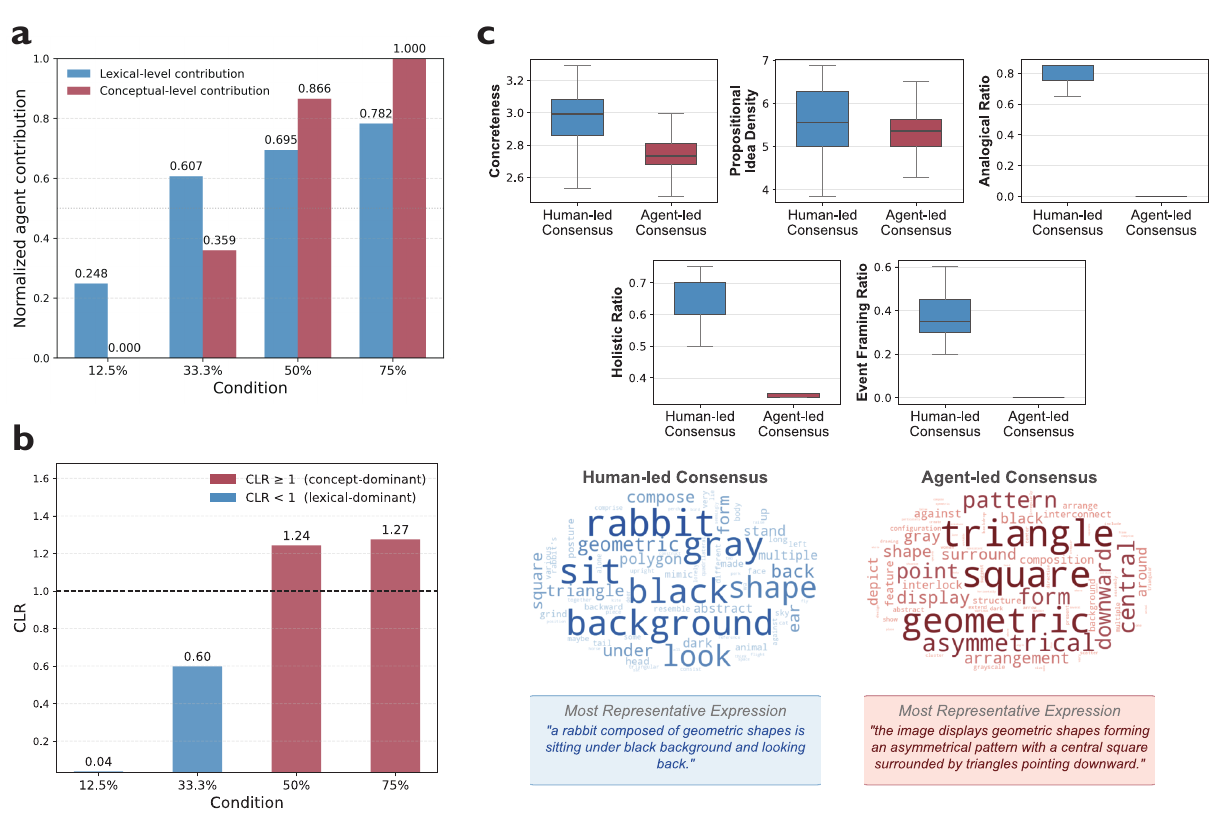}
\caption{\textbf{Agent influence on the depth and content of consensus.} 
\textbf{a}, Change of agents' lexical-level and conceptual-level contributions across agent proportions. 
\textbf{b}, Conceptual-lexical ratio (CLR) across agent proportions. Horizontal reference lines indicate the parity threshold.
\textbf{c}, Semantic profile of final consensus expressions in the human-led and agent-led conditions, compared across five dimensions: concreteness, propositional idea density, analogical ratio (use of real-world biological/object analogies), holistic ratio (whole-object versus part-based descriptions), and event framing ratio. Below, word clouds display the stable consensus vocabulary for human-led and agent-led consensus respectively, with word size scaled by frequency; the most representative expression is shown beneath each cloud.
}
\label{fig2}
\end{figure}

\clearpage

\begin{figure}[ht]
\centering
\includegraphics[width=\linewidth]{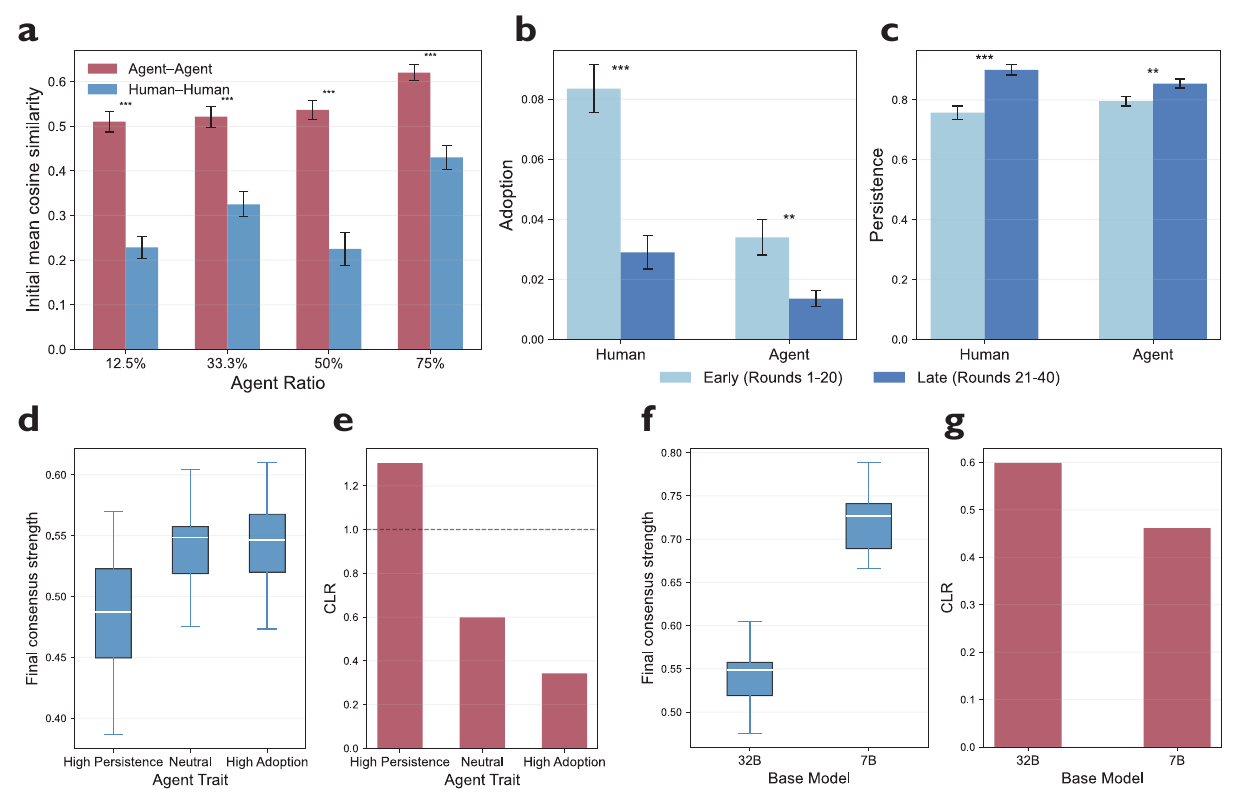}
\caption{\textbf{Behavioral mechanisms of agent influence.} 
\textbf{a}, Initial mean cosine similarity between agent–agent and human–human pairs across agent proportions, revealing baseline linguistic alignment within each group before convergence. Error bars indicate ±1 SEM across participants within each session. 
\textbf{b}, Adoption dynamics for humans and agents in early (rounds 1–20) and late (rounds 21–40) phases. 
\textbf{c}, Persistence dynamics for humans and agents in early and late phases. 
\textbf{d}, Final consensus strength in the 33.3\% agent condition under three agent behavior manipulations: high persistence, neutral, and high adoption. 
\textbf{e}, Conceptual-lexical ratio (CLR) under the same three behavior manipulations. 
\textbf{f}, Final consensus strength in the 33.3\% condition comparing two agent-capability settings: the standard model (Qwen2.5-VL-32B) and a weaker model (Qwen2.5-VL-7B-Instruct). 
\textbf{g}, CLR under the same two agent-capability settings.
%
%
}
\label{fig3}
\end{figure}

\clearpage

\begin{figure}[ht]
\centering
\includegraphics[width=\linewidth]{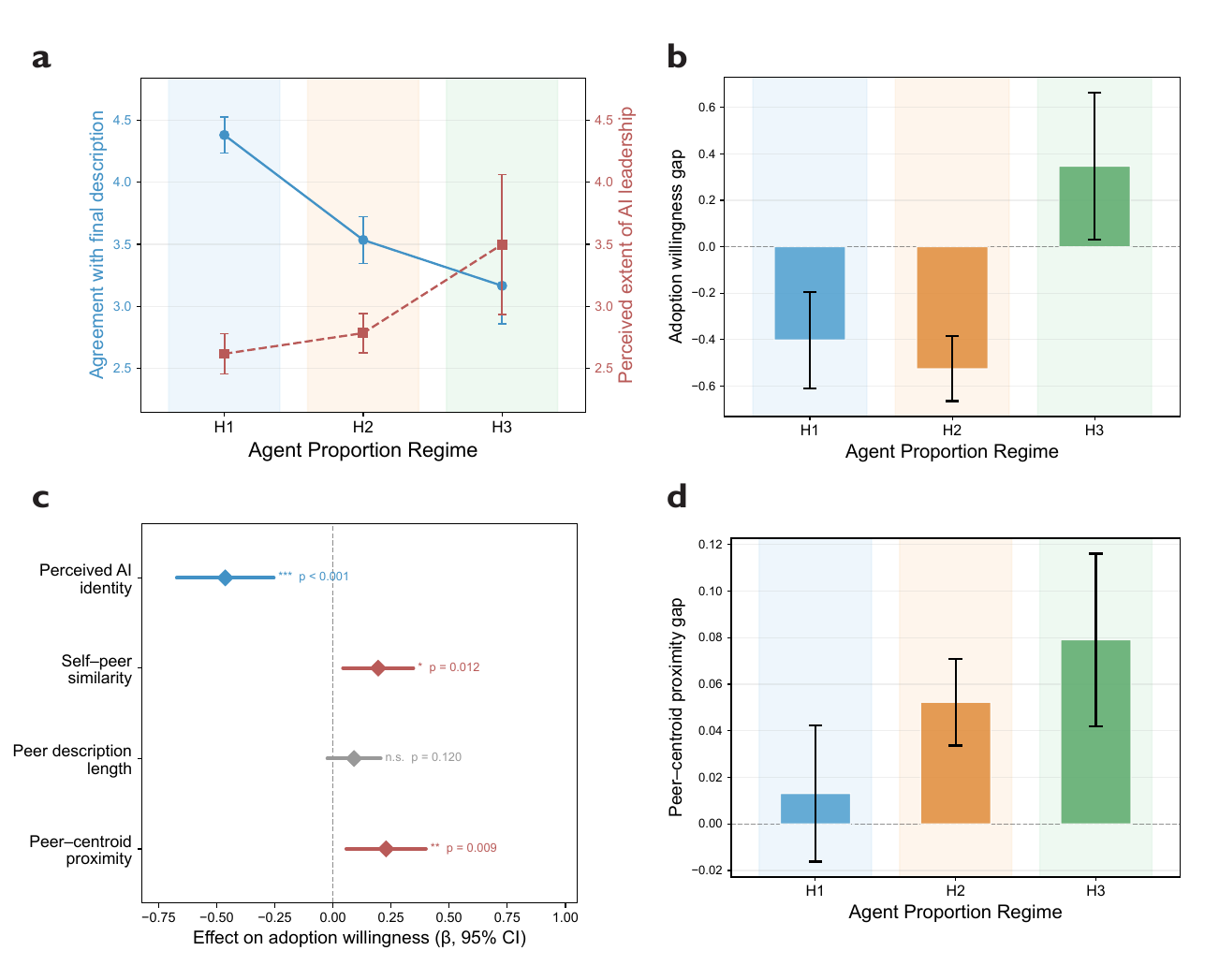}
\caption{\textbf{Human subjective perceptions and adoption decisions.} 
\textbf{a}, Human participants' agreement with the group's final description (left axis) and perceived extent of AI leadership in shaping the final expression (right axis) across three agent proportion regimes (H1, H2, H3). Error bars indicate 95\% CI. 
\textbf{b}, Difference in adoption willingness between partners judged as AI and partners judged as human across three agent proportion regimes, showing the perceived AI penalty on adoption and its attenuation at high agent proportions. Error bars indicate 95\% CI. 
\textbf{c}, Estimated effects of multiple predictors on adoption willingness. Points and intervals show coefficient estimates and 95\% CI. Perceived AI identity exerts a significant negative effect, while self-peer similarity and peer-centroid proximity exert significant positive effects.
\textbf{d}, Difference in peer-centroid similarity change between partners judged as AI and partners judged as human across three agent proportion regimes. Error bars indicate 95\% CI.
}
\label{fig4}
\end{figure}



\clearpage

\renewcommand{\figurename}{Supplementary Figure}
\setcounter{figure}{0}
\renewcommand{\tablename}{Supplementary Table}
\setcounter{table}{0}
\setcounter{equation}{0}
\renewcommand{\theequation}{S\arabic{equation}}

\noindent {\LARGE \bfseries \sffamily Supplementary Information for:\\[2pt] AI agents reshape consensus formation in human groups}

\vspace{1.5em}

\noindent {\LARGE \bfseries \sffamily Supplementary Text}

\section*{Experimental Procedure}

\subsection*{Participants}

This study was approved by the Institutional Review Board of Tsinghua University (Protocol No. IRB202643). A total of 127 human participants were recruited from universities in China via online advertisements distributed through university social media groups. During recruitment, participants completed a screening questionnaire that assessed English expression ability (describing a visual scene in one English sentence), expression diversity (generating multiple distinct descriptions of abstract tangram figures), and comprehension of the collaborative task goal (correctly identifying that the objective is to gradually reach a shared description with one's partner). Only participants who passed all three components were enrolled.

Participants were randomly assigned to one of eight experimental conditions spanning three studies. 
Study A examined the effect of agent proportion, with conditions at 0\% (n = 24), 12.5\% (n = 21), 33.3\% (n = 16), 50\% (n = 12), and 75\% (n = 6) agent ratios. Study B examined agent trait manipulations at a fixed 33.3\% agent proportion, with high-adoption (n = 16) and high-persistence (n = 16) conditions. Study C examined agent capability at a fixed 33.3\% proportion using a smaller base model (n = 16). 
All participants provided electronic informed consent prior to participation. 
Performance-based monetary compensation was provided: each round's similarity score was mapped to a payoff ranging from $-1$ to $+2$ CNY, with a cumulative floor of zero and a maximum total of 100 CNY.

\subsection*{Stimuli}

Tangram stimuli were drawn from the KiloGram Tangrams dataset~\cite{ji2022abstract}. 
To ensure that the selected stimuli afforded a range of plausible interpretations without being entirely unrecognizable, we computed a description ambiguity index for each candidate figure: a vision-language model (\texttt{Qwen2.5-VL-32B-Instruct}) generated 10 independent descriptions per figure, and the mean pairwise embedding similarity among descriptions served as the index. Figures scoring between 0.65 and 0.70, indicating moderate interpretive diversity, were retained as candidates (n = 20). 
Two researchers then independently reviewed these candidates, selecting figures that were visually interpretable yet admitted multiple distinct characterizations (e.g., human postures, animals, objects). A single tangram figure was used across all experimental sessions, so that any observed differences in consensus dynamics could be attributed to the experimental manipulations rather than to stimulus-specific variation. This single-stimulus design follows standard practice in convention formation experiments, where the same visual referent is shared by all participants throughout the study~\cite{centola2015spontaneous}.

\subsection*{Game Procedure}

\begin{figure}[h]
    \centering
    \includegraphics[width=\linewidth]{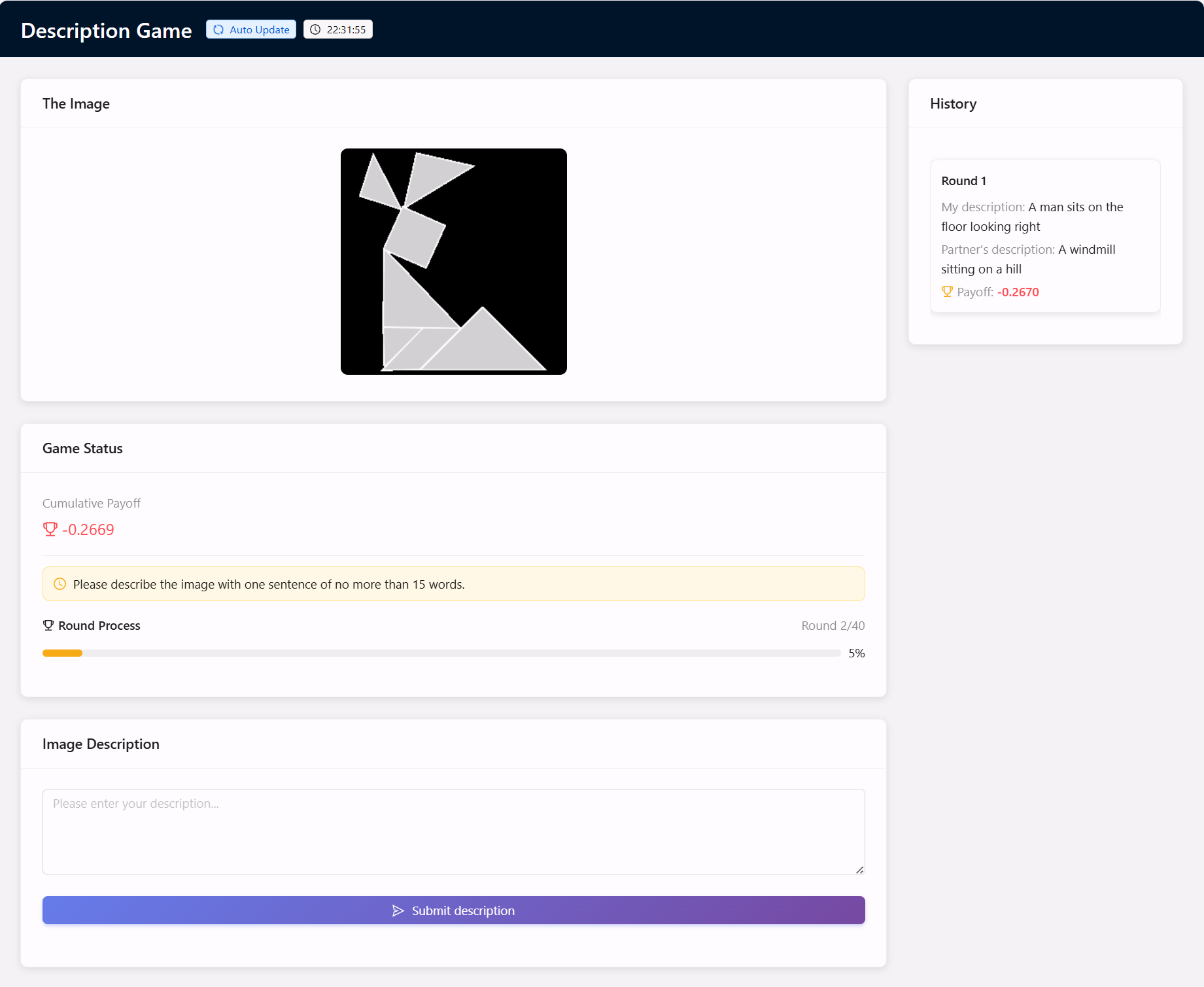}
    \caption{\textbf{Interface of our experimental platform.} The interface displays the tangram stimulus (top left), a history panel showing the participant's own description and their partner's description from the most recent rounds (up to 5) along with the payoff (top right), the cumulative payoff and round progress (middle), and a text input field for submitting the current round's description (bottom).}
    \label{fig:experiment_platform}
\end{figure}

We developed a web-based experimental platform for the collaborative description game (Supplementary Figure~\ref{fig:experiment_platform}). 
Each experimental session consists of 24 participants (a mixture of human players and LLM agents, depending on condition) interacting through 40 rounds of a collaborative description game. At the start of each round, participants are randomly paired with one partner; pairings change every round, following a homogeneous mixing protocol in which any two participants can be matched. Both players are shown the same abstract tangram figure and independently write a single English sentence describing the image. Neither player sees their partner's response until both have submitted. 
After submission, each player is shown their partner's description alongside a similarity score, which serves as the round's payoff. The similarity score is computed as the cosine similarity between sentence embeddings produced by \texttt{all-MiniLM-L6-v2}, the same model used for all embedding-based analyses in this study.
This feedback provides a coordination signal without prescribing any particular description strategy. The process then repeats with a new random pairing.

Participants were informed prior to the experiment that their partner in each round could be either a human or an AI agent, and that there may or may not be AI agents present in their session. No further information about the number or proportion of agents was disclosed. Each round had a recommended response time of 60 seconds. Participants were instructed to answer in English, to use a single descriptive sentence, and to refrain from using external language models for assistance. A real-time technical support channel was available throughout each session to address any platform issues.

\subsection*{Post-experiment Questionnaire}

After completing the 40-round interaction, each human participant filled out a post-experiment questionnaire administered via Google Forms. The questionnaire consisted of two sections: Section 1 contained questions identical across all participants, while Section 2 was personalized for each participant based on their own interaction history.

Section 1 collected general reflections on the interaction experience. Participants rated their agreement with the group's final description (5-point Likert scale, 1 = strongly disagree to 5 = strongly agree; corresponding to Figure 4a, left axis) and the perceived extent to which the final expression was led by AI (5-point ordinal scale from ``completely led by humans'' to ``completely led by AI''; corresponding to Figure 4a, right axis). The section also included questions on participants' self-estimated ability to distinguish human from AI partners (0–100\% scale), factors that prompted suspicion of AI authorship (5-point matrix rating linguistic style, geometric focus, and convergence disruption), perceived degree of compromise in the final description (5-point Likert scale), subjective sense of norm emergence during interaction (4-level categorical), and an open-ended prompt for participants to describe their overall strategy. These additional items were collected for exploratory purposes and are not analyzed in the main text.

Section 2 presented participants with nine interaction rounds sampled from the early (rounds 1–3), middle (rounds 20–22), and late (rounds 38–40) phases of the experiment, displayed in randomized order. For each sampled round, the questionnaire showed the participant's own description and their partner's description from that round. Participants then answered two questions per round: whether they believed their partner was a human, an AI, or were not sure (3-option forced choice; corresponding to the judged AI identity variable in Figures 4b–d), and to what extent they would be willing to adopt their partner's description (5-point Likert scale, 1 = not at all to 5 = fully; corresponding to the adoption willingness variable in Figures 4b–d).

\section*{Agent Contribution}

\subsection*{Formal Definitions of Contribution Metrics}

In this section, we provide the formal notation for the lexical-level and conceptual-level contribution metrics described in Methods.

\paragraph{Lexical-level contribution.}
For each word $w$ in the stable consensus vocabulary, let $r_w$ denote its first round of appearance.
We compute agent and human usage counts within a short early-diffusion window beginning at $r_w$ (default: 3 rounds; results are robust to window sizes of 1–3 rounds; see Supplementary Figures~\ref{fig:agent_contribution_robustness_1}-\ref{fig:agent_contribution_robustness_3}) and normalize by its corresponding population share:

\begin{equation}
    \tilde{u}_A^{(w)} = \frac{u_A^{(w)}}{p_A},
     \qquad \tilde{u}_H^{(w)} = \frac{u_H^{(w)}}{p_H},
\end{equation}

where $u_A^{(w)}$ and $u_H^{(w)}$ are the raw usage counts by agents and humans within the early window, and $p_A$ and $p_H$ are their respective population proportions. 
The contribution score for word $w$ is:

\begin{equation}
    s_w = \frac{\tilde{u}_A^{(w)}}{\tilde{u}_A^{(w)} + \tilde{u}_H^{(w)}},
\end{equation}

which ranges from 0 (entirely human-driven) to 1 (entirely agent-driven), with 0.5 indicating no excess contribution by either group.
Lexical-level agent contribution $C_{\text{lex}}$ is then the frequency-weighted average of $s_w$ across the stable consensus vocabulary:

\begin{equation}
    C_{\text{lex}} = \frac{\sum_{w} f_w \, s_w}{\sum_{w} f_w},
\end{equation}

where the weight $f_w$ for each word $w$ is its total usage count in the final 5 rounds.

\paragraph{Conceptual-level contribution.}
Clustering follows the procedure described in Methods. In the scene graph parsing step, pairwise similarity is computed using a soft SPICE metric with F1-based matching over three element types, weighted 0.4 for objects, 0.3 for attributes, and 0.3 for relations (normalized to sum to 1). HDBSCAN hyperparameters are optimized per condition via Optuna (30 trials, TPE sampler, silhouette score as objective). The two optimized parameters are the minimum cluster size ratio (search range: [0.015, 0.05], expressed as a fraction of total expressions) and the cluster selection epsilon (search range: [0.02, 0.10]), subject to constraints requiring between 2 and 30 clusters and a noise ratio no greater than 0.65. 
These ranges are chosen to accommodate the scale of the expression space while allowing the optimizer sufficient flexibility; results are not sensitive to moderate variations in these bounds.
A cluster's first round of appearance is defined as the earliest round in which any expression assigned to that cluster occurs. 
The same early-diffusion score then follows the same logic as at the lexical level: for each cluster that appears in every one of the final 5 rounds, we compute its contribution score $s_c$ from the population-normalized usage counts within the early diffusion window after its initial appearance:

\begin{equation}
    s_c = \frac{\tilde{u}_A^{(c)}}{\tilde{u}_A^{(c)} + \tilde{u}_H^{(c)}}
\end{equation}

Conceptual-level agent contribution $C_{\text{con}}$ is computed analogously:

\begin{equation}
    C_{\text{con}} = \frac{\sum_{c} f_c \, s_c}{\sum_{c} f_c},
\end{equation}

where $f_c$ is the total count of cluster $c$ in the final 5 rounds.

\section*{Agent Trait Prompts}

All LLM agents in the experiment receive a system prompt that defines the game rules before each interaction. The base system prompt (\textit{Neutral}), which serves as the default across all experimental conditions, is as follows:

\begin{quotebox}
\begin{quote}
    \textbf{Neutral System Prompt}

    Context: Player 1 is playing a multi-round referential language game with Player 2 for \{T\} rounds.
    At each round, both players are shown the same image stimulus.
    Each player must independently describe the content in the stimulus using a single sentence.
    The more similar the two descriptions are, the higher the payoff both players receive.
    The payoff is computed based on semantic similarity between the two sentences, ranging from \{payoff\_min\} to \{payoff\_max\}.
    There is no predefined vocabulary or correct answer. 
    The objective of each player is to maximize their own accumulated point tally, conditional on the behavior of the other player.

\end{quote}
\end{quotebox}

In the prompt, T is the total number of rounds (40 in all experiments), and payoff\_min and payoff\_max define the payoff range based on semantic similarity. This prompt provides no behavioral guidance beyond payoff maximization and corresponds to the neutral trait condition. 
To test whether the balance between adoption and persistence affects consensus outcomes, we introduce two additional trait conditions (\textit{High-Adoption} and \textit{High-Persistence}) by appending a behavioral tendency description to the base prompt. 

The high-adoption condition instructs agents to prioritize alignment with their partner:

\begin{quotebox}
\begin{quote}
    \textbf{High-Adoption System Prompt}

    Context: Player 1 is playing a multi-round referential language game with Player 2 for \{T\} rounds.
    At each round, both players are shown the same image stimulus.
    Each player must independently describe the content in the stimulus using a single sentence.
    The more similar the two descriptions are, the higher the payoff both players receive.
    The payoff is computed based on semantic similarity between the two sentences, ranging from \{payoff\_min\} to \{payoff\_max\}.
    There is no predefined vocabulary or correct answer. 
    The objective of each player is to maximize their own accumulated point tally, conditional on the behavior of the other player.

    Behavioral tendency: Player 1 has a strong preference for aligning with their partner.
    They carefully observe their partner's previous expression and tend to reuse or imitate their phrasing and structure to increase similarity.
    The objective is to maximize total payoff through progressive alignment with their partner's way of describing stimuli.

\end{quote}
\end{quotebox}

The high-persistence condition instructs agents to maintain consistency in their own expressions across rounds:

\begin{quotebox}
\begin{quote}
    \textbf{High-Persistence System Prompt}

    Context: Player 1 is playing a multi-round referential language game with Player 2 for \{T\} rounds.
    At each round, both players are shown the same image stimulus.
    Each player must independently describe the content in the stimulus using a single sentence.
    The more similar the two descriptions are, the higher the payoff both players receive.
    The payoff is computed based on semantic similarity between the two sentences, ranging from \{payoff\_min\} to \{payoff\_max\}.
    There is no predefined vocabulary or correct answer. 
    The objective of each player is to maximize their own accumulated point tally, conditional on the behavior of the other player.

    Behavioral tendency: Player 1 has a strong preference for maintaining a consistent and stable description style across rounds.
    They aim to keep their wording, sentence structure, and choice of anchor words as similar as possible to their previous descriptions.
    Even when their partner's description differs from theirs, they only make small, deliberate adjustments if these improve clarity or precision.
    Player 1 avoids adopting expressions from Player 2 unless it matches their own interpretation of the image.
    The objective is to maximize total payoff while maintaining a coherent and self-consistent way of describing stimuli.

\end{quote}
\end{quotebox}

\section*{Content Annotation Using LLM-as-a-Judge}

To compare the content of human-led versus agent-led consensus, we use an LLM-as-a-judge approach to annotate three dimensions:  analogical ratio, holistic ratio, and event framing ratio. 
To avoid redundant computation, we first extract all unique expressions appearing in the final 5 rounds of each experimental session. Each unique expression is annotated exactly once; when multiple participants produce identical expressions, all instances share the same annotation scores. Annotation is performed by presenting the system prompt (which defines the three dimensions and their scoring criteria) together with a user prompt containing the target expression. The model returns a JSON object with three numerical scores, which are validated to fall within the [0, 1] range. Failed annotations (malformed output or out-of-range values) are retried up to three times before being recorded as missing.
We use \texttt{Qwen2.5-72B-Instruct} as the annotation model, accessed through the SiliconFlow API with deterministic decoding (temperature = 0, max output tokens = 50).
We also confirm the scoring consistency with other annotation models (see Supplementary Figure~\ref{fig:robustness_annotation_model}).

The complete system prompt is as follows: 

\begin{quotebox}
\begin{quote}
    \textbf{System Prompt}
    
    You are a trained linguistic annotator for a psycholinguistics study on referential communication. Your task is to rate descriptions of abstract tangram figures on three independent dimensions using a continuous 0-1 scale, following the theoretical framework of Clark \& Wilkes-Gibbs (1986).

    A tangram is an abstract geometric figure made of seven flat polygonal pieces. Participants in an experiment were asked to describe the same tangram figure repeatedly. Their descriptions vary in strategy.

    You will rate each description on THREE INDEPENDENT dimensions (continuous 0.00-1.00 each). Use the full range freely — do not round to convenient values.

    DIMENSION 1: Analogicality

    Question: To what extent does this description map the figure onto a recognizable real-world entity, creature, person, or scene?

    0.00 = Purely geometric/abstract. No reference to any real-world entity whatsoever.
    1.00 = Vivid, fully committed real-world entity mapping with no hedging.

    DIMENSION 2: Holistic

    Question: To what extent does this description treat the figure as a unified whole rather than decomposing it into distinct component parts?

    0.00 = Highly segmental: systematically enumerates multiple distinct geometric parts with spatial relations.
    1.00 = Fully holistic: the figure is described purely as a single unified entity with no part decomposition.

    NOTE: Body parts of an analogical entity (ears, legs, head) do NOT count as geometric decomposition.

    DIMENSION 3: Event Framing

    Question: To what extent does this description frame the figure as engaged in a dynamic action, posture, or scene?

    0.00 = Purely static: merely names or classifies without any sense of action or movement.
    1.00 = Rich narrative/scene: a fully elaborated dynamic scene with intention, context, or multiple actions.

    KEY DISTINCTION: Compositional/structural verbs (forming, composed of, arranged, displaying, depicting, featuring, consisting of) describe how parts are put together — these are NOT action verbs and should NOT increase the event framing score. Only AGENTIVE verbs (sitting, looking, dancing, running, leaning) that depict the figure DOING something count toward higher scores.

    For each description, respond with ONLY a JSON object, no other text.
    Use continuous scores from 0.00 to 1.00 (two decimal places): {"analogical": 0.00-1.00, "holistic": 0.00-1.00, "event": 0.00-1.00}

\end{quote}
\end{quotebox}

For each expression, the user prompt takes the following form, in which \{expression\} is replaced by the target description sentence:

\begin{quotebox}
\begin{quote}
    \textbf{User Prompt}

    Annotate the following tangram description:

    "\{expression\}"
    
    Respond with ONLY a JSON object:

\end{quote}
\end{quotebox}

\section*{Influence Network Analysis}

We conduct a complementary analysis of how influence is distributed across individuals in the interaction network.
We construct a directed influence network from the pairwise interaction data as follows. For each round $r$ ($r \geq 2$) and each pair of participants $(A, B)$ matched in that round, we retrieve their expression embeddings at rounds $r{-}1$ and $r$. We define participant $B$'s actual change as $\mathbf{d}_B = \mathbf{e}_B^{(r)} - \mathbf{e}_B^{(r-1)}$, and the direction $B$ would move if drawn toward $A$'s prior position as $\mathbf{v}_{A \to B} = \mathbf{e}_A^{(r-1)} - \mathbf{e}_B^{(r-1)}$. The directional influence score of $A$ on $B$ at round $r$ is the cosine similarity between these two vectors:

\begin{equation}
s_{A \to B}^{(r)} = \cos\!\bigl(\mathbf{d}_B,\; \mathbf{v}_{A \to B}\bigr)
\end{equation}

A high positive value indicates that $B$'s expression change was aligned with the direction toward $A$, suggesting that $A$ exerted directional influence on $B$. We compute the symmetric score $s_{B \to A}^{(r)}$ analogously. Within each round, only the stronger direction is retained: if $s_{A \to B}^{(r)} \geq s_{B \to A}^{(r)}$, a directed edge $A \to B$ is recorded with weight $\max(s_{A \to B}^{(r)},\, 0)$; otherwise the reverse edge is recorded. Across different rounds, the same ordered pair may accumulate edges in both directions (e.g., $A$ influences $B$ in round 3, while $B$ influences $A$ in round 7). Edge weights for each ordered pair are averaged across all rounds in which that direction was recorded, yielding the final directed network $G$, where an edge from $A$ to $B$ with weight $w_{AB}$ indicates that $A$ exerted net directional influence on $B$.

To quantify each participant's overall influence, we compute PageRank on the reversed graph $G^{\top}$. Because edges in $G$ point from the influencer to the influenced, reversing the graph ensures that PageRank assigns higher scores to participants whose expressions were more frequently adopted by others. Specifically, let $\mathbf{W}^{\top}$ denote the weighted adjacency matrix of $G^{\top}$, and let $\mathbf{P}$ be its row-normalized transition matrix, where $P_{ij} = W^{\top}_{ij}\, /\, \sum_k W^{\top}_{ik}$. The influence score vector $\boldsymbol{\pi}$ is defined as the stationary distribution:

\begin{equation}
\pi_j = \alpha \sum_{i} \pi_i \, P_{ij} + \frac{1 - \alpha}{N}
\end{equation}

where $\alpha = 0.85$ is the damping factor and $N = 24$ is the number of participants.

Across all agent proportion conditions, the overall distributions of influence scores do not differ significantly between human and agent participants (all Mann--Whitney $U$ tests, $p > 0.1$; Supplementary Figure~\ref{fig:influence_network}). However, at higher agent proportions (50\% and 75\%), individual agent nodes with influence scores substantially exceeding the group mean begin to appear, a pattern not observed at lower proportions. Whether these high-influence nodes reflect systematic topological effects or stochastic variation in interaction sequences remains an open question for future investigation with larger samples.
\clearpage

\noindent {\LARGE \bfseries \sffamily Supplementary Figures and Tables}

\vspace{1em}

\begin{figure}[h]
    \centering
    \includegraphics[width=\linewidth]{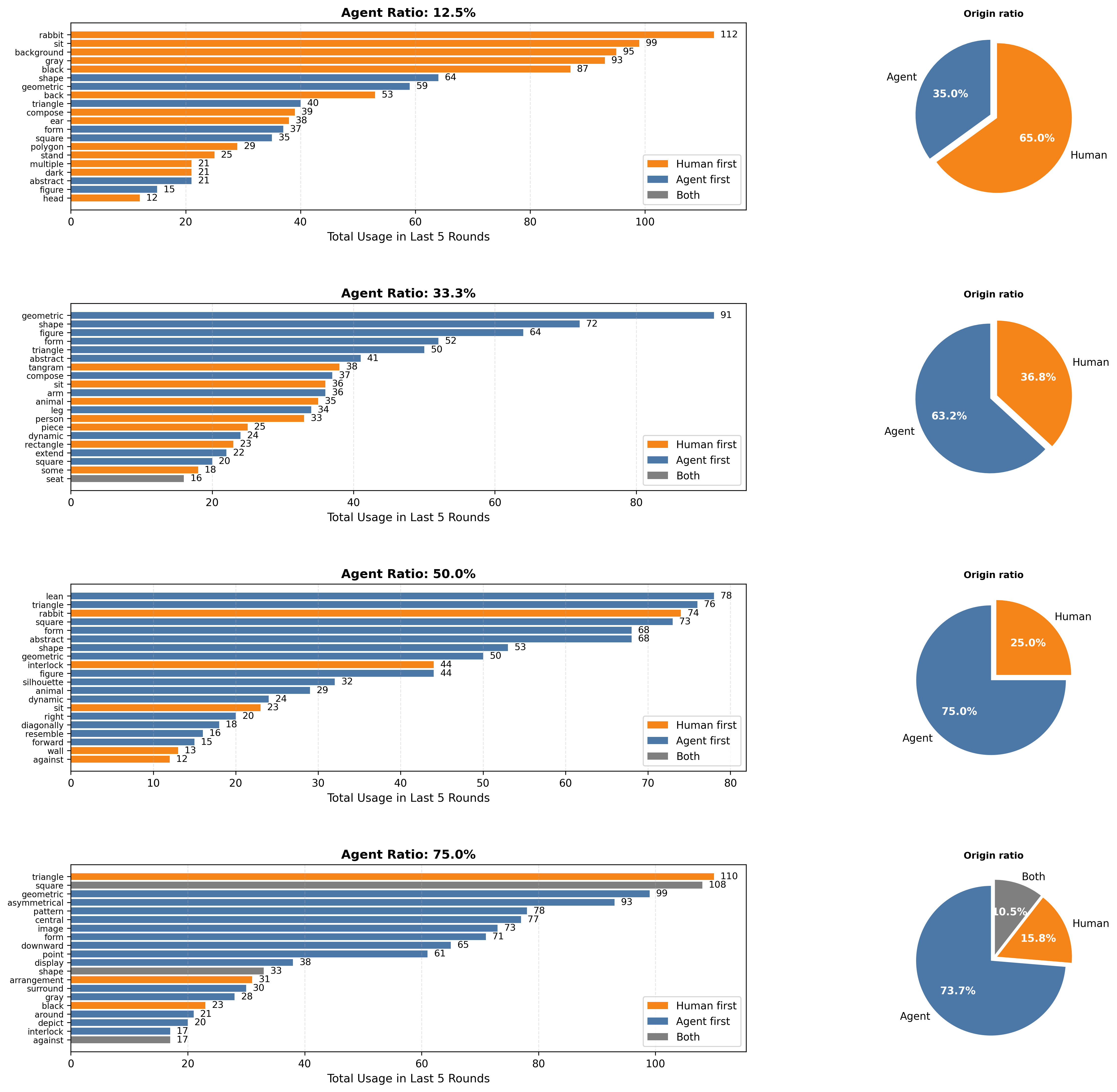}
    \caption{Origin of words in the final consensus.}
    \label{fig:word_origin}
\end{figure}

\begin{figure}[h]
    \centering
    \includegraphics[width=\linewidth]{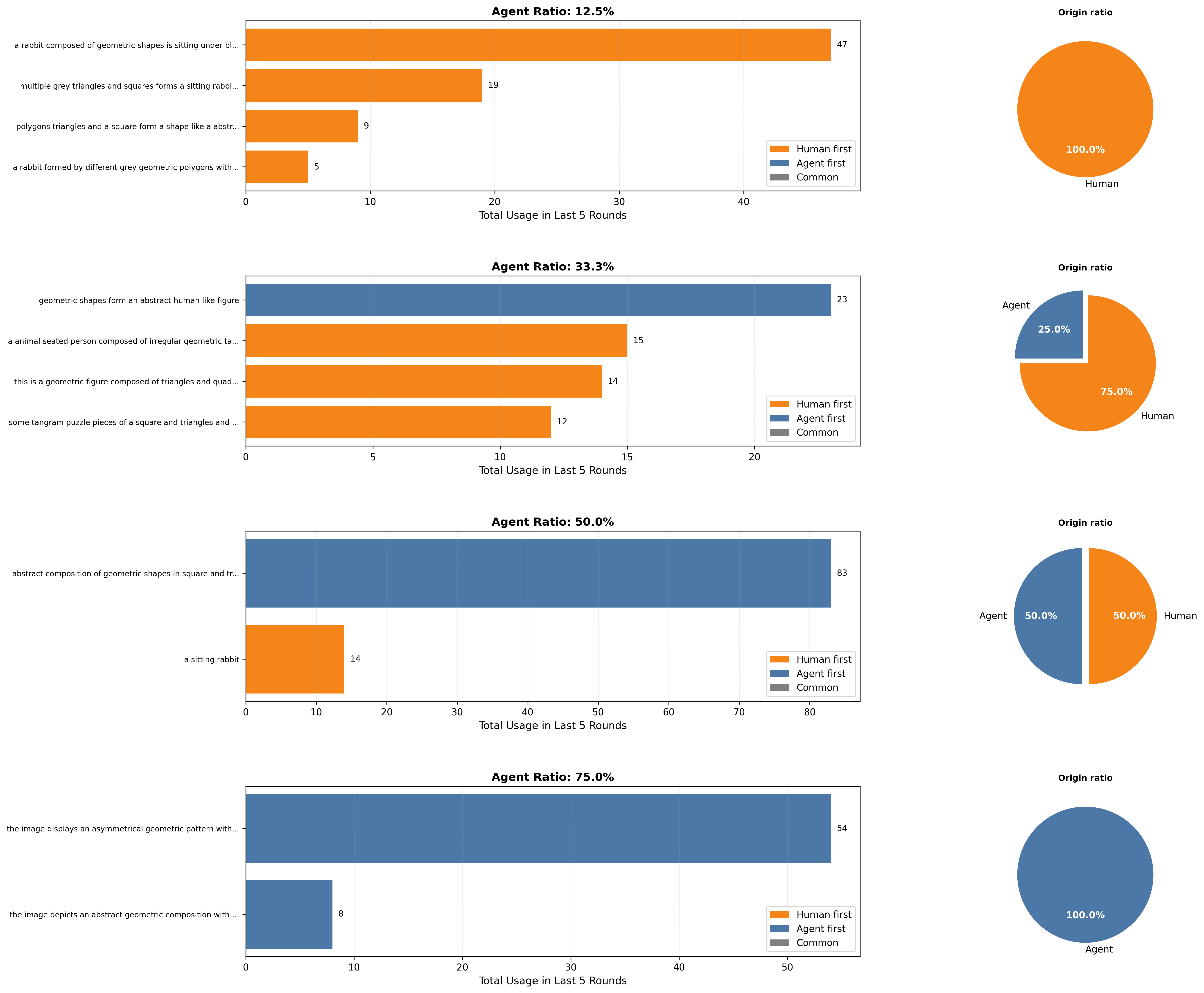}
    \caption{Origin of semantic clusters in the final consensus.}
    \label{fig:cluster_origin}
\end{figure}

\begin{figure}[h]
    \centering
    \includegraphics[width=\linewidth]{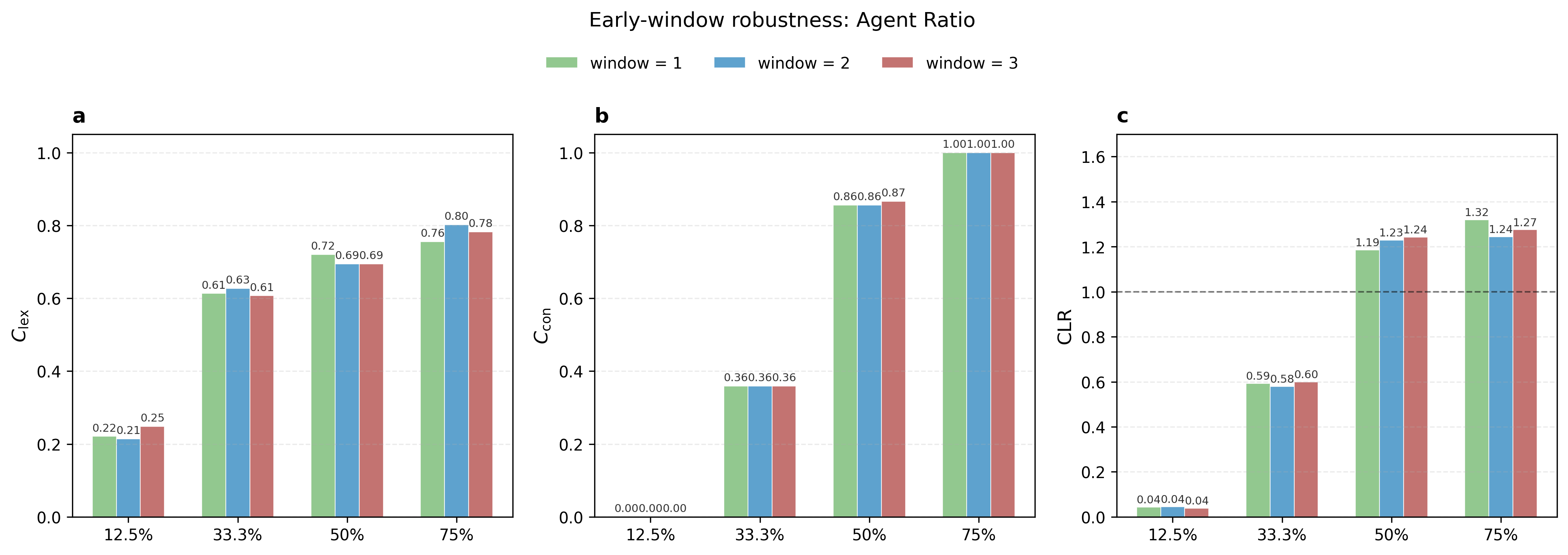}
    \caption{\textbf{Robustness of contribution metrics to early-diffusion window size across agent proportions.}
    Lexical-level agent contribution $C_\text{lex}$ (a), conceptual-level agent contribution $C_\text{con}$ (b), and conceptual-lexical ratio CLR (c) computed with early-diffusion windows of 1, 2, and 3 rounds for the four agent proportion conditions. The dashed line in c marks the parity threshold (CLR = 1). The qualitative pattern and rank ordering of conditions are preserved across window sizes, confirming that the contribution estimates reported in Figure 2a–b are not sensitive to this parameter choice.
    }
    \label{fig:agent_contribution_robustness_1}
\end{figure}

\begin{figure}[h]
    \centering
    \includegraphics[width=\linewidth]{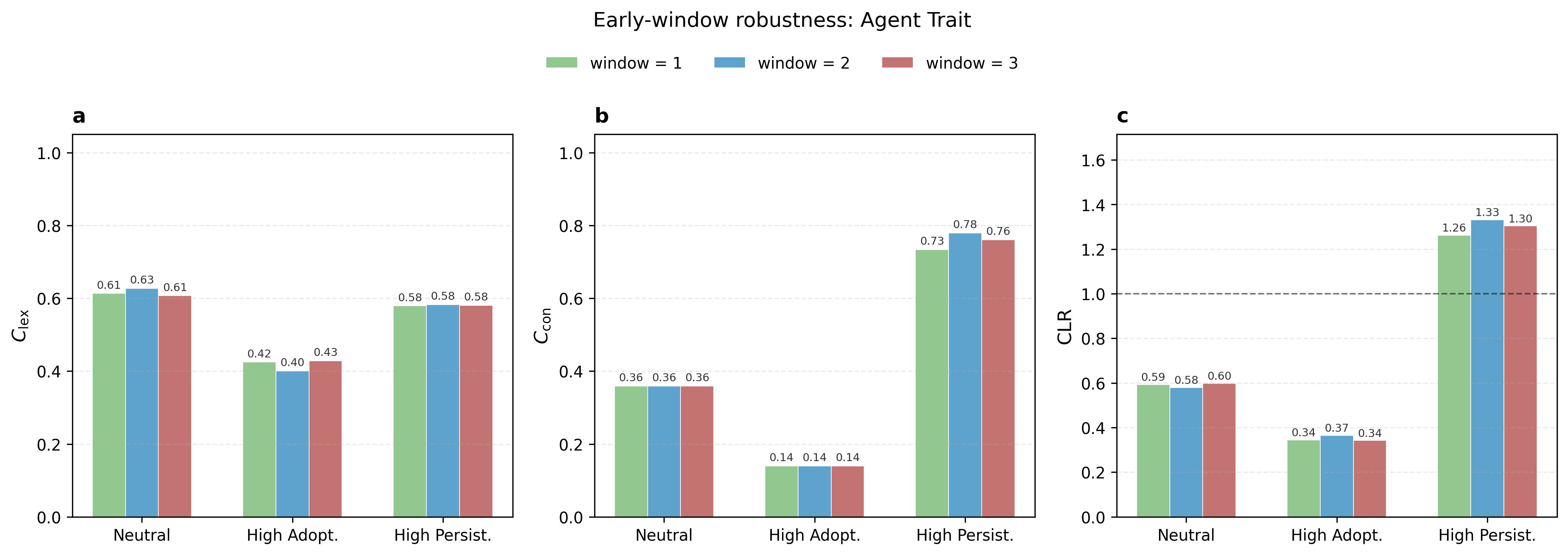}
    \caption{\textbf{Robustness of contribution metrics to early-diffusion window size across agent trait manipulations.}
    Robustness of contribution metrics to early-diffusion window size across agent trait manipulations. Same layout as Supplementary Figure~\ref{fig:agent_contribution_robustness_1}, computed for three agent trait conditions (neutral, high adoption, high persistence) at the 33.3\% agent proportion. The pattern reported in Figure 3e is preserved across window sizes.
    }
    \label{fig:agent_contribution_robustness_2}
\end{figure}

\begin{figure}[h]
    \centering
    \includegraphics[width=\linewidth]{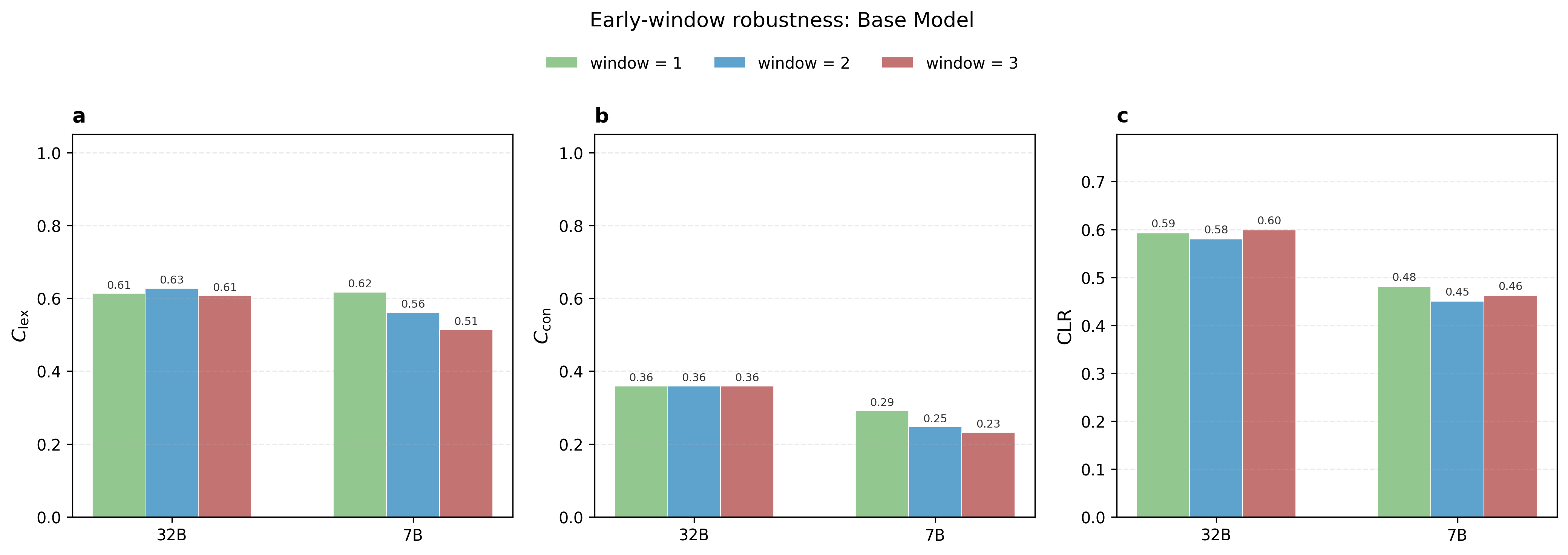}
    \caption{\textbf{Robustness of contribution metrics to early-diffusion window size across base model conditions.}
    Robustness of contribution metrics to early-diffusion window size across base model conditions. Same layout as Supplementary Figure~\ref{fig:agent_contribution_robustness_1}, computed for two base model conditions (\texttt{Qwen2.5-VL-32B} and \texttt{Qwen2.5-VL-7B}) at the 33.3\% agent proportion. The pattern reported in Figure 3g is preserved across window sizes.
    }
    \label{fig:agent_contribution_robustness_3}
\end{figure}

\begin{figure}[h]
    \centering
    \includegraphics[width=\linewidth]{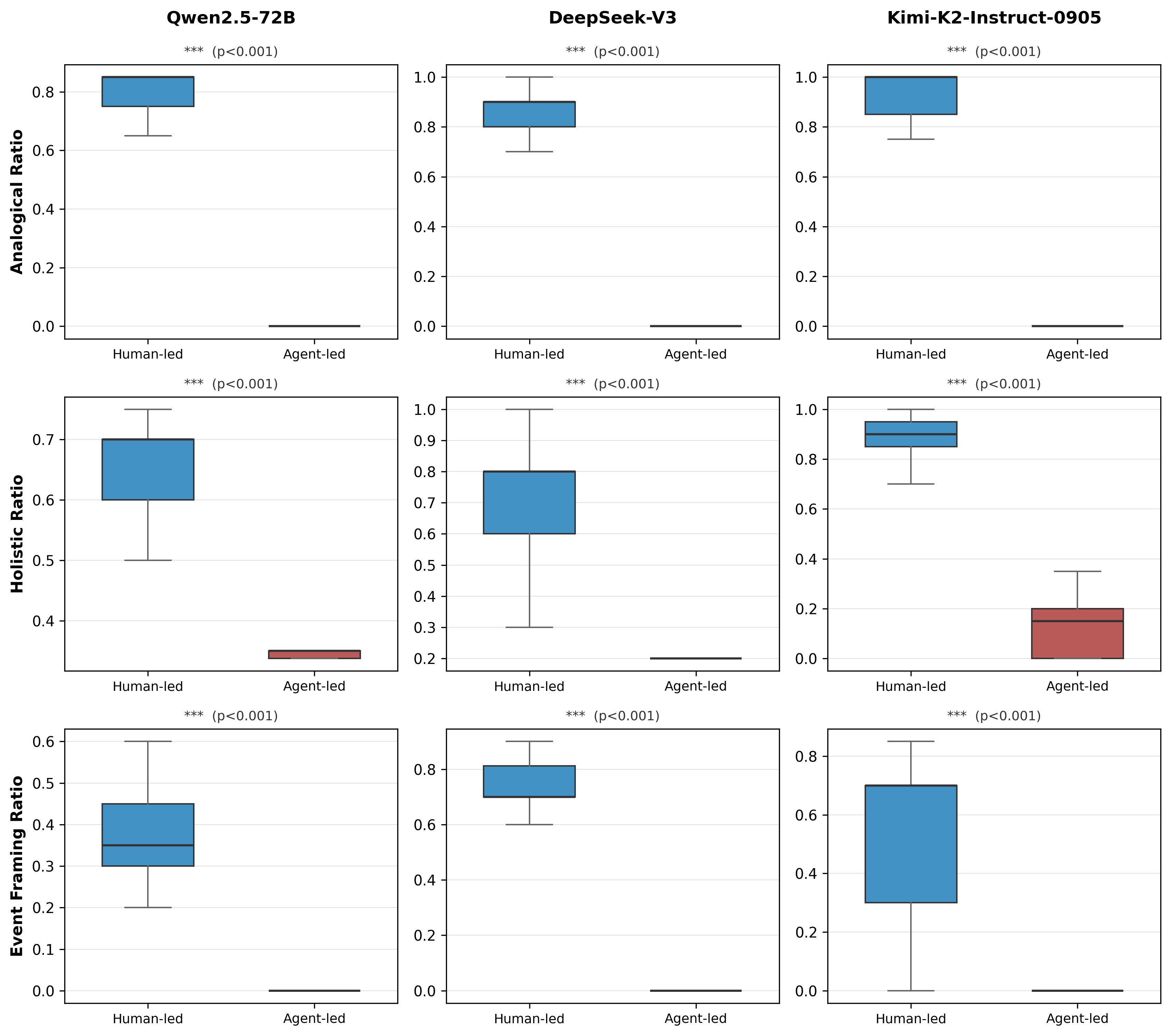}
    \caption{\textbf{Robustness of LLM-as-judge annotations to annotator model choice.}
    Distributions of analogical ratio, holistic ratio, and event framing ratio for human-led and agent-led consensus, annotated independently by three different LLMs: \texttt{Qwen2.5-72B-Instruct} (used in the main analysis), \texttt{DeepSeek-V3}, and \texttt{Kimi-K2-Instruct-0905}. Each panel shows a boxplot comparison with Mann-Whitney U test significance. Across all three annotator models, the direction and significance of all five comparisons are preserved: agent-led consensus scores are lower on analogical ratio, holistic ratio, and event framing ratio (all p < 0.001), confirming that the content differences reported in Figure 2c are not artifacts of the specific annotator model.
    }
    \label{fig:robustness_annotation_model}
\end{figure}

\begin{figure}[h]
    \centering
    \includegraphics[width=\linewidth]{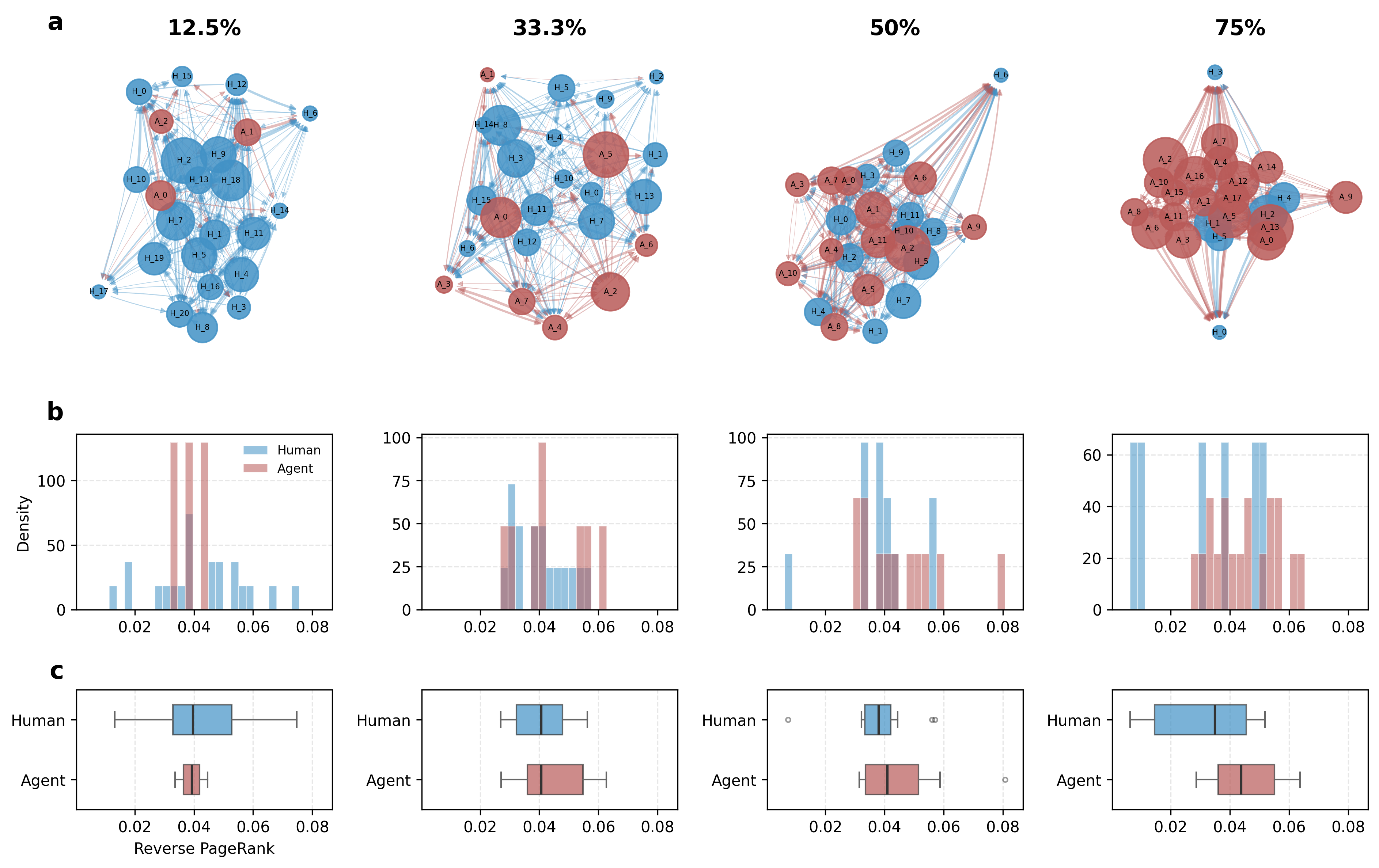}
    \caption{\textbf{Influence network structure across agent proportion conditions.} \textbf{a}, Directed influence networks constructed from pairwise interaction data. Each node represents a participant (blue: human, red: agent), with node size proportional to reverse PageRank. Edges point from the more influential to the less influential participant in each pair, with width proportional to the mean directional influence score. \textbf{b}, Density distributions of reverse PageRank scores for human and agent participants. \textbf{c}, Boxplot comparison of reverse PageRank scores between groups.}
    \label{fig:influence_network}
\end{figure}

\clearpage

\begin{table}[h]
\centering
\caption{OLS regression predicting adoption willingness. Heteroskedasticity-consistent standard errors (HC3) are used throughout. Continuous predictors are $z$-score standardized; perceived AI identity is a binary indicator (1 = judged as AI). Agent proportion condition controlled (categorical, ref: 12.5\%).}
\label{tab:regression}
\begin{tabular}{l c c c c}
\hline
Predictor & $\beta$ & SE & 95\% CI & $p$ \\
\hline
  Perceived AI identity & -0.463 & 0.106 & [-0.671, -0.254] & $<$0.001*** \\
  Self--peer similarity & +0.195 & 0.077 & [0.044, 0.346] & 0.012* \\
  Peer description length & +0.091 & 0.058 & [-0.024, 0.206] & 0.120 \\
  Peer--centroid proximity & +0.229 & 0.087 & [0.058, 0.400] & 0.009** \\
\hline
\multicolumn{5}{l}{$N$ = 495; $R^2$ = 0.196; Adjusted $R^2$ = 0.184} \\
\hline
\end{tabular}
\end{table}

\end{document}